\documentclass[conference]{IEEEtran}
\IEEEoverridecommandlockouts
\usepackage{url}
\usepackage{cite}
\usepackage{hyperref}
\usepackage{amsthm,amsmath,amssymb,amsfonts}
\usepackage{array}
\usepackage{graphicx}
\usepackage{textcomp}
\usepackage{xcolor}
\usepackage[ruled]{algorithm2e}
\usepackage{algpseudocode}
\usepackage{subfigure}
\usepackage{multirow}
\usepackage{pifont}
\usepackage{booktabs}
\usepackage{float}
\definecolor{mypurple}{RGB}{153,51,255}
\definecolor{myblue}{RGB}{0,102,255}
\usepackage{soul} 
\usepackage{color, xcolor} 

\DeclareRobustCommand*{\IEEEauthorrefmark}[1]{%
    \raisebox{0pt}[0pt][0pt]{\textsuperscript{\footnotesize\ensuremath{#1}}}}
\begin{document}
\title{Structure-Preserving Graph Representation Learning }

\author{\IEEEauthorblockN{Ruiyi Fang\IEEEauthorrefmark{1},
Liangjian Wen\IEEEauthorrefmark{2}\thanks{§ Ruiyi Fang and Liangjian Wen have equal contributions.},
Zhao Kang\IEEEauthorrefmark{2}\thanks{* Zhao Kang is the corresponding author.}, 
Jianzhuang Liu\IEEEauthorrefmark{3}}
\IEEEauthorblockA{\IEEEauthorrefmark{1}School of Information and Software Engineering, 
 University of Electronic Science and Technology of China}
\IEEEauthorblockA{\IEEEauthorrefmark{2}School of Computer Science and Engineering, 
 University of Electronic Science and Technology of China}
\IEEEauthorblockA{\IEEEauthorrefmark{3}Shenzhen Institutes of Advanced Technology,
University of Chinese Academy of Sciences}

 \ \{fangruiyi8, wlj68164\}@gmail.com, zkang@uestc.edu.cn, jz.liu@siat.ac.cn}





\maketitle

\begin{abstract}

Though graph representation learning (GRL) has made significant progress, it is still a challenge to extract and embed the rich topological structure and feature information in an adequate way. Most existing methods focus on local structure and fail to fully incorporate the global topological structure. To this end, we propose a novel Structure-Preserving Graph Representation Learning (SPGRL) method, to fully capture the structure information of graphs. Specifically, to reduce the uncertainty and misinformation of the original graph, we construct a feature graph as a complementary view via  $k$-Nearest Neighbor method.
The feature graph can be used to contrast at node-level  to capture the local relation. Besides, we retain the global topological structure information by maximizing the mutual information (MI) of the whole graph and feature embeddings, which is theoretically reduced to exchanging the feature embeddings of the feature and the original graphs to reconstruct themselves.
Extensive experiments show that our method has quite superior performance on semi-supervised node classification task and excellent robustness under noise perturbation on graph structure or node features. The source code is available at \url{https://github.com/uestc-lese/SPGRL}.

\end{abstract}







\section{Introduction}
Ubiquitous graph or network data expressed in the form of node connections and features raise a new challenge for traditional machine learning techniques to discover knowledge \cite{lin2021multi}. Graph convolutional network (GCN) has proved to be a powerful tool to handle graph-structured data in a variety of domains, such as social network~\cite{qiu2018deepinf}, chemistry~\cite{duvenaud2015convolutional}, biology~\cite{rhee2018hybrid}, traffic prediction~\cite{cui2019traffic}, text classification~\cite{hamilton2017inductive}, and knowledge graph~\cite{wang2018cross}. Most GCN-based methods learn a low-dimensional and dense representation by reconstructing the feature or graph in the autoencoder framework~\cite{wang2016structural,cao2016deep,liu2022multilayer}. 
How to fully inherit the rich information from topological structure and node attribute is crucial to the success of GCN~\cite{kipf2017semi}.


\begin{figure*}[!htbp]
  \centering
  \includegraphics[width=0.8\textwidth]{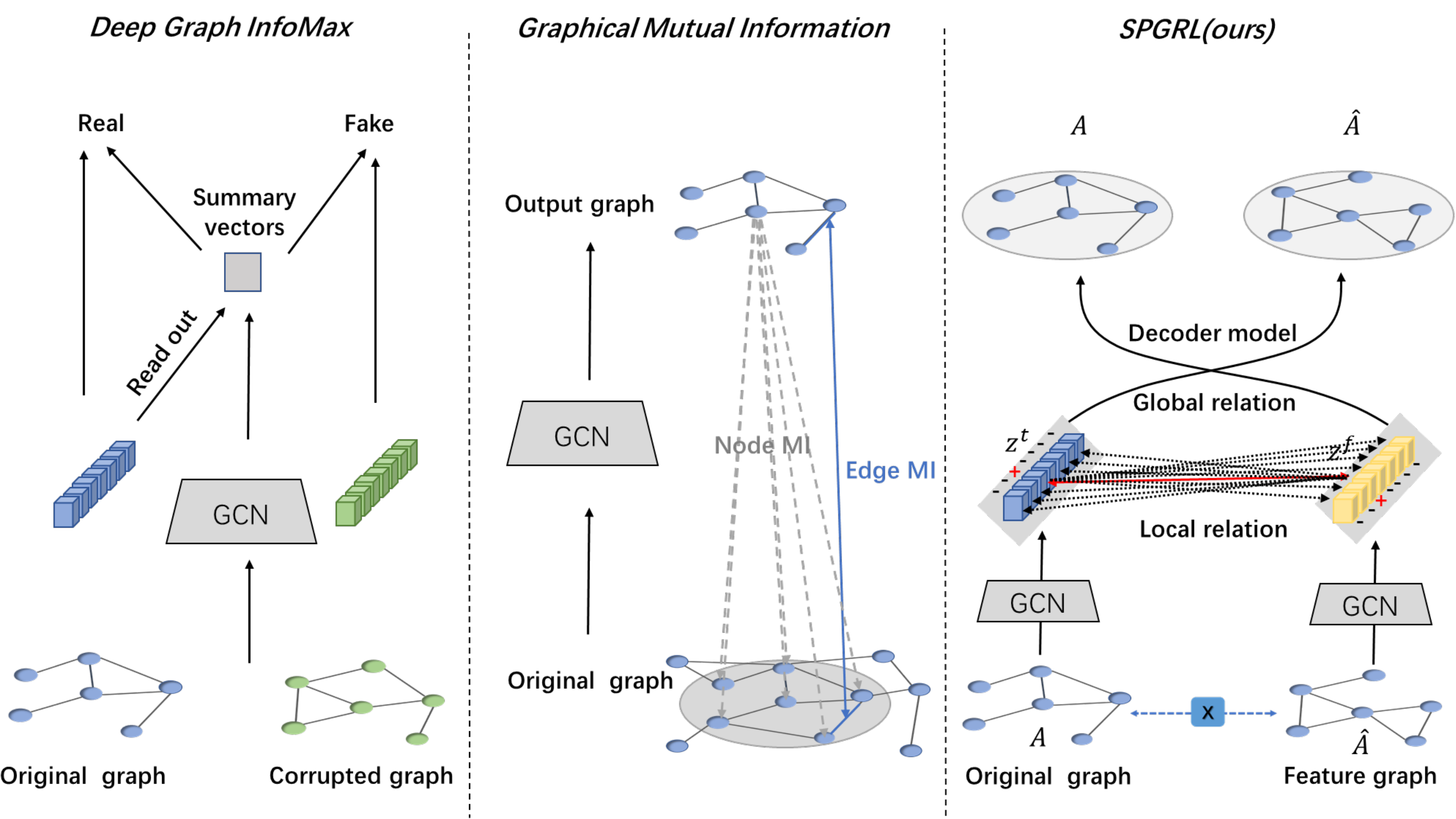}
  \caption{An overview of DGI (left), GMI (middle) and SPGRL (right). Different from them, SPGRL maximizes the MI between the graph and feature embedding to explore the global structure information. }
  \label{fig:overview}
\end{figure*}

Basically, GCN processes graph by means of aggregating features from neighborhood nodes~\cite{li2018deeper}. In essence, it performs as low-pass filtering on feature vectors of nodes and
graph structure only provides a way to denoise the data~\cite{DBLP:journals/corr/abs-1905-09550, wu2019simplifying}. 
Some works have theoretically analyzed the weaknesses of GCN in feature information fusion~\cite{wang2020gcn}. Unlike some other deep neural networks, stacking multiple layers leads to over-smoothing, which seriously degrades the feature discriminability and deteriorates the performance of downstream tasks~\cite{ma2020towards}.

In order to better fuse the feature information, graph attention network (GAT)~\cite{velickovic2018graph} has been proposed, which can assign an adaptive weight to each edge of the graph. Later, Wang \emph{et al.}~\cite{wang2020gcn} propose adaptive 
multi-channel GCN (AMGCN), which better fuses the topological structure and feature information through the attention mechanism.
However, the attention-based approach needs to calculate the weight of each edge, which consumes much computation time and memory for large graphs. 

Recently, contrastive learning, as a burgeoning unsupervised learning mechanism, has achieved superior performance in various tasks \cite{pan2021multi}. It learns effective representations by contrasting positive samples against negative samples through the design of pretext tasks including the design of data augmentation schemes and object functions. For instance, GRACE~\cite{Zhu:2020vf} maximizes the agreement of node representation in two views constructed by data augmentation strategy. SLAPS~\cite{fatemi2021slaps} solves the problem of underutilization of information in unsupervised learning by constructing a homogeneous node graph and contrasting it. GCA~\cite{zhu2021graph} proposes an adaptive data augmentation scheme to preserve the intrinsic structure and properties of the graph by exploiting the connection patterns of original graph. These methods mainly explore the local relation without preserving structural information. Recently, maximizing mutual information (MI)  has been adopted to explore rich information from topological structure and node features \cite{tsai2021self}. Deep Graph InfoMax (DGI)~\cite{veli2018deep} maximizes the MI between the hidden representation and a summary vector. However, its simple averaging readout function damages the distinguish capability between nodes and makes the global-level representation unreliable. 
These methods largely rely on "augmentation engineering", which requires extensive domain knowledge and even incurs negative effects.

To get rid of above issue, some other methods use two neural networks to learn from each other to boost performance. For example, SCRL~\cite{liu2021self} performs representation consistency constraint by constructing feature graph and topology graph for cross-prediction, and effectively improves the feature information fusion ability of GCN. Some other data augmentation strategies, such as GEN~\cite{wang2021graph} and PTDNet~\cite{luo2021learning}, have also been developed. Graphical Mutual Information (GMI)~\cite{peng2020graph} instead tries to maximize the MI between the target node and its neighbors at node-level, and the proximity topological structure at the edge level. As shown in Fig.\ref{fig:overview}, maximizing MI at multiple levels does not really consider MI at the global level. They mainly align embeddings between the same nodes in different topological structures, rather than using the local-global relationships. 
To better explore the global structure, we propose a novel Structure-Preserving Graph Representation Learning (SPGRL) method, which maximizes the MI between topology graph and feature embeddings. First, we construct a feature graph to provide a complementary view, which allows the feature information to propagate through the feature space, thus reinforcing the feature information and alleviating the uncertainty or error in the original graph.
The original graph is often extracted from complex interaction systems that inevitably involve uncertain, redundant, wrong and missing connections~\cite{wang2021graph}. 
Specifically, we use $k$-Nearest Neighbor ($k$NN) method to build the feature graph $\mathbf{\hat{A}}$, which could preserve high-order proximity. Then, the output embedding $\mathbf{Z^t}$ from $A$ and $\mathbf{Z^f}$ from $\mathbf{\hat{A}}$ are obtained through GCN. They are refined by local node-level relation through contrastive loss. Finally, we maximize MI between embeddings and topology graph, which is theoretically equivalent to minimizing exchange reconstruction loss.
Therefore, we reconstruct original graph with the embedding of feature graph and reconstruct feature graph with the embedding of original graph.
Our main contributions are summarized as follows:
\begin{itemize}
    \item We propose to preserve the global structure information by maximizing the MI between topology graph and feature embeddings. Theoretical analysis shows that this can be achieved by exchange reconstruction.
   \item Our method explores the local node-level relation with the aid of feature graph. Feature view preserves high-order relations and helps eliminate the uncertainty or error in the original graph. 
    \item Comprehensive experiments on benchmarks show the superior performance of our method compared to other state-of-the-art methods in semi-supervised node classification task. Our method also outperforms other mainstream methods even with very few labels and under noise perturbation.
\end{itemize}

\section{RELATED WORKS}
\subsection{Graph Representation Learning}
In the last decades, a large number of methods have been proposed to learn the representation of graph data.
Most early GRL methods are based on random walk. Inspired by the Skip-gram model used for natural language processing, Perozzi \emph{et al.} propose DeepWalk~\cite{perozzi2014deepwalk}, which learns latent representations by utilizing local information obtained from truncated random walks. Subsequently, several variants have been proposed to improve DeepWalk, prominent examples include LINE~\cite{tang2015line} and node2vec~\cite{grover2016node2vec}.
Due to the success of deep learning, graph neural network (GNN) approach has been developed. ChebNet~\cite{defferrard2016convolutional} uses the Chebyshev polynomial approximation to optimize a general graph convolutional framework based on graph Laplacian. GCN~\cite{kipf2017semi} further simplifies the convolution operation using a localized first-order approximation. GAT~\cite{velickovic2018graph} assigns different attention weights to different nodes in the neighborhood to better fuse node features.
Demo-Net~\cite{wu2019net} builds a degree-specific GNN for the representation of nodes and graphs. MixHop~\cite{abu2019mixhop} utilizes multiple powers of adjacency matrix to learn general mixing of neighborhood information.
However, these methods only use a single topology graph for node aggregation. Some methods propose to solve this problem for better fusing node features by constructing feature graph. AMGCN~\cite{wang2020gcn} utilizes attention mechanism to merge embeddings extracted from topology graph and feature graph. However, these attention-based approaches are often computation expensive. 

\subsection{Self-supervised Learning}
The role of self-supervised learning is to learn representative representations without label information, which can reduce the human cost of annotating data. Many successful applications of self-supervised learning have emerged, from natural language processing~\cite{lample2019cross} to computer vision~\cite{pathak2016context}.
Contrastive learning, a class of self-supervised learning, trains networks by comparing representations learned from augmented samples.
For instance, MoCo~\cite{he2020momentum} and SimCLR~\cite{chen2020simple} construct negative and positive sample pairs by data augmentation techniques and then contrast their embeddings. 
However, it is computationally expensive for large datasets. Another category is cluster-based approaches. For example, Caron \emph{et al.}~\cite{caron2020unsupervised} proposes a simplified training pipeline that maps features to cluster prototypes. 
Recently, there have been several works focusing on self-supervised learning methods in the graph domain.
M3S~\cite{sun2020multi} utilizes a multi-stage, self-supervised learning approach to improve the generalization performance of GCN.
GRACE~\cite{Zhu:2020vf} is a graph contrastive representation learning framework that seeks an optimal common representation. 
GCA ~\cite{zhu2021graph} uses adaptive graph structure augmentation to construct a contrastive view and distinguishes the embeddings of the same node in two different views from the embedding of other nodes.
SLAPS~\cite{fatemi2021slaps} solves the problem of underutilization of information in unsupervised learning by constructing a homogeneous node graph at graph level and contrasting it. 
DGI~\cite{veli2018deep} first proposes the use of MI in the graph domain, which maximizes MI between hidden representation and a summary vector from a corrupted graph. But DGI's simple averaging readout function compromises global information.
Unlike them, GMI~\cite{peng2020graph} uses a discriminator to directly measure the MI between the input graph and output graph in terms of features and edges, not directly using local-global relationships. Due to these design flaws, they fail to take full advantage of the global graph information.
Furthermore, most contrastive learning methods involve random destruction at nodes and edges. This could introduce noise to the original graph data and reduce the generalizability of the learned representations. 
Hence, there is much room to improve information utilization at the node level and graph level.

\section{The Proposed Methodology}

The aim of our proposed method is to fully exploit potential correlations between graph structure and node attributes. In particular, not just capturing graph information from the original graph, we also exploit the feature view via feature graph. Ultimately we inherit rich representation information from feature graph view and topology graph view by maximizing global level MI.

\subsection{Feature Extraction}
We first outline the general setting of graph representation learning. A graph can be represented as $\mathbf{G}=\left\{ \mathbf{A}, \mathbf{X}\right\}$, 
where $\mathbf{A} \in \mathbb{\mathbf{R}}^{\mathbf{N}\times \mathbf{N}}$ is the adjacency matrix of $\mathbf{N}$ nodes 
and $\mathbf{X \in \mathbb{R}^{N \times d}}$ is the node feature matrix, i.e., 
each node is described by a vector with $\mathbf{d}$ dimensions and belongs to one out of $\mathbf{M}$ classes.
$\mathbf{A_{ij}}=1$ represents that there is an edge between node $\mathbf{i}$ and $\mathbf{j}$, otherwise
$\mathbf{A_{ij}}=0$. In our study, we derive the feature graph $\mathbf{\hat{G}=\left\{ \hat{A}, X\right\}}$,
which shares the same $\mathbf{X}$ with $\mathbf{G}$, but has a different adjacency matrix.
Therefore, topology graph and feature graph refer to 
$\mathbf{G}$ and $\mathbf{\hat{G}}$ respectively.


To represent the structure of nodes in the feature space, we build feature graph $\mathbf{\hat{G}}$ via $k$NN. First, a similarity matrix \textbf{$S$} is computed using the Cosine similarity :
\begin{equation}
    S_{ij} = \frac{\mathbf{x_{i}}\cdot \mathbf{x_{j}}}{\|\mathbf{x_{i}}\| \cdot \|\mathbf{x_{j}}\|}
    \label{cos},
\end{equation}
where $\mathbf{S_{ij}}$ is the similarity between node feature $\mathbf{x_i}$ and node feature $\mathbf{x_j}$.
Then, for each node, we choose the top $k$ nearest neighbors and
establish edges. In this way, we construct the structure of the feature graph as $\hat A$.

To extract meaningful features from graph, we adopt GCN as our backbone. 
With the input graph $\mathbf{G}$, the ${(l+1)}$-th layer's output $H^{(l+1)}$ can be represented as:
\begin{equation}
    H^{(l+1)} = ReLU(D^{-\frac{1}{2}}AD^{-\frac{1}{2}}H^{(l)}W^{(l)}).
\end{equation}
where $ReLU$ is the Relu activation function ($ReLU(\cdot) = max(0, \cdot) $), 
$\mathbf{D}$ is the degree matrix of $\mathbf{A}$, $W^{(l)}$ is a layer-specific trainable weight matrix, 
$H^{(l)}$ is the output matrix in the $l$-th layer and $H^{(0)} = X$.
In our study, we use two GCNs to exploit the information in topology and feature space. 
The output is denoted by $\mathbf{Z^t}=\left\{ \mathbf{z^t_{1}}, \mathbf{z^t_{2}}, \cdots, \mathbf{z^t_{N}}\right\}$ 
and $\mathbf{Z^f}=\left\{ \mathbf{z^f_{1}}, \mathbf{z^f_{2}}, \cdots, \mathbf{z^f_{N}}\right\}$, respectively.


\subsection{Local Node-level Relation}

Unlike previous graph contrastive learning models, SPGRL uses feature graph as a complementary view to capture local relation at the node-level. The feature graph characterizes high-order relations, thus the feature view encodes high-order structure information. Therefore, it provides complementary information to the original graph, which just describes the first-order relation and inevitably involves uncertainty or error. To learn a consistent representation, we uncover the local pairwise relations between nodes via a contrastive learning mechanism. Concretely, we treat $\mathbf{z^t_{i}}$ as a positive sample of $\mathbf{z^f_{j}}$ only when $\mathbf{i=j}$ satisfies and $\mathbf{z^t_{i}}$ are negative samples of $\mathbf{z^f_{j}}$ for $\mathbf{i \neq j}$, and vice versa. Then the loss can be formulated as:

\begin{equation}
\centering
\begin{aligned}
  {L}_{cr}= &-\sum_{i=1}^{N}\log \frac{\exp(sim(\mathbf{z^t_{i}},\mathbf{z^f_{i}}))}{\exp(sim(\mathbf{z^t_{i}},\mathbf{z^f_{i}})) + \sum_{\mathbf{j=1, j \neq i}}^{N}\exp(sim(\mathbf{z^t_{i}},\mathbf{z^f_{j}}))} \\
                  &- \sum_{i=1}^{N}\log \frac{\exp(sim(\mathbf{z^f_{i}},\mathbf{z^t_{i}}))}{\exp(sim(\mathbf{z^f_{i}},\mathbf{z^t_{i}})) + \sum_{\mathbf{j=1,j \neq i}}^{N}\exp(sim(\mathbf{z^f_{i}},\mathbf{z^t_{j}}))},
\label{equ: cr_loss}
\end{aligned}
\end{equation} 

where $sim(\cdot,\cdot)$ is the cosine function as defined in Eq.(\ref{cos}). Intuitively, the purpose of Eq.(\ref{equ: cr_loss}) is to make the representations of nodes within local neighborhood as close as possible and the representations of nodes from different groups as distinct as possible. 

\subsection{Global Graph-level Relation}
Node contrastive method is not an effective way to attain global structural information in the topology graph. Existing
approaches ignore the mutual corroboration effects of structures
and attributes. The embedding of feature graph is expected to extract some relevant structure information 
from topology graph to improve the accuracy of downstream tasks. To this end,
we propose to maximize the MI $I(\mathbf{Z^f},\mathbf{A})$ between $\mathbf{Z^f}$ and whole topology graph $\mathbf{A}$ to preserve the structure information in topology graph. In addition, we also improve the embedding of topology graph $\mathbf{Z^t}$ by maximizing $I(\mathbf{Z^t},\mathbf{\hat{A}})$ between $\mathbf{Z^t}$  and whole feature graph $\mathbf{\hat{A}}$.


Let's take $I(\mathbf{Z^f},\mathbf{A})$ as example to show the computation process. Mathematically,
\begin{align}
 I(\mathbf{Z^f},\mathbf{A}) = \mathbb{E}_{p(\mathbf{Z^f}, \mathbf{A})}\left[\log \frac{p(\mathbf{Z^f}, \mathbf{A})}{p(\mathbf{Z^f})p(\mathbf{A})}\right].
\end{align}
According to the relation between entropy and MI, we can decompose $ I(\mathbf{Z^f},\mathbf{A})$ as follows:
\begin{align}
 I(\mathbf{Z^f},\mathbf{A}) =H(\mathbf{A})-H(\mathbf{A}|\mathbf{Z^f}),
\end{align}
where $H(\mathbf{A}|\mathbf{Z^f})=-\mathbb{E}_{p(\mathbf{Z^f}, \mathbf{A})}\left[\log p(\mathbf{A}|\mathbf{Z^f}  )\right]$ is the conditional entropy, and $H(\mathbf{A})$, the entropy of $\mathbf{A}$, is irrelevant to $\mathbf{Z^f}$. Hence, maximizing $I(\mathbf{Z^f},\mathbf{A})$ is equivalent to maximizing $-H(\mathbf{A}|\mathbf{Z^f})$.
However, the computation of $H(\mathbf{A}|\mathbf{Z^f})$ is intractable due to unknown of the condition distribution $p(\mathbf{A}|\mathbf{Z^f} )$. 

We assume $q_{\phi}(\mathbf{A}|\mathbf{Z^f} )$ is a variational approximation to $p(\mathbf{A}|\mathbf{Z^f} )$. Since $KL(p(\mathbf{A}|\mathbf{Z^f} )||q_{\phi}(\mathbf{A}|\mathbf{Z^f} ))	\geq 0 $, we can derive that: 
\begin{align}
\mathbb{E}_{p(\mathbf{Z^f}, \mathbf{A})}\left[\log p(\mathbf{A}|\mathbf{Z^f}  )\right]   \geq \mathbb{E}_{p(\mathbf{Z^f}, \mathbf{A})}\left[\log q_{\phi}(\mathbf{A}|\mathbf{Z^f}  )\right].
\end{align}
Hence, $\mathbb{E}_{p(\mathbf{Z^f}, \mathbf{A})}\left[\log q_{\phi}(\mathbf{A}|\mathbf{Z^f}  )\right]$ is the lower bound of \\ $\mathbb{E}_{p(\mathbf{Z^f}, \mathbf{A})}\left[\log p(\mathbf{A}|\mathbf{Z^f}  )\right]$.
Specifically, $q_{\phi}(\mathbf{A}|\mathbf{Z^f} )$ can be regarded as the decoder function whose equation is as follows:
\begin{align}
q_{\phi}(\mathbf{A}|\mathbf{Z^f})=\prod_{i=1}^{N} \prod_{j=1}^{N} q_{\phi}\left(\mathbf{A_{i j}} \mid \mathbf{z^f_{i}}, \mathbf{z^f_{j}}\right),
\end{align}
where the probability of an edge existing between two nodes is:
\begin{align}
q_{\phi}\left(\mathbf{A_{i j}}=1 \mid \mathbf{z^f_{i}}, \mathbf{z^f_{j}}\right)=\operatorname{sigmoid}\left(\mathbf{z^{f_{i}^{T}}}\mathbf{z^f_{j}}\right).
\end{align}

Above optimization objective of maximizing $I(\mathbf{Z^f},\mathbf{A})$ is equivalent to:
\begin{align}
L_{re}^{\mathbf{A}}=E_{p(\mathbf{Z^f}, \mathbf{A})}[\log q_{\phi}(\mathbf{A} \mid \mathbf{Z^f})].
\end{align}
Likewise, we can obtain a similar objective of maximizing $I(\mathbf{Z^t},\mathbf{\hat{A}})$ as follows: 
\begin{align}
L_{re}^{\mathbf{\hat{A}}}=E_{p(\mathbf{Z^t}, \mathbf{\hat{A}})}[\log q_{\phi}(\mathbf{\hat{A}} \mid \mathbf{Z^t})].
\end{align}
To summarize, we propose the exchange-reconstruction mechanism to maximize $I(\mathbf{Z^f},\mathbf{A})$ and $I(\mathbf{Z^t},\mathbf{\hat{A}})$ between the embeddings and graph structures.
Then the global MI loss can be formulated as:
\begin{equation}
\centering
\begin{aligned}
  {L}_{re}= L_{re}^{\mathbf{A}} + L_{re}^{\mathbf{\hat{A}}}.
\label{equ: re_loss}
\end{aligned}
\end{equation}

\subsection{Node Classification}

Ideally, $\mathbf{Z^t}$ and $\mathbf{Z^f}$ should be close to each other.
To preserve the information from feature graph and topology graph, $\mathbf{Z^t}$ and $\mathbf{Z^f}$ are concatenated as the consensus representation $R$~\cite{liu2021self}.
Then we use $\mathbf{R}$ for semi-supervised classification, which is realized through a linear transformation and a softmax function.
$\mathbf{B}$ and $\mathbf{a}$ are weights and bias of the linear layer, respectively. $\mathbf{Y^{\prime}}$ is the prediction result and $\mathbf{Y^{\prime}_{ij}}$ is the probability 
of node $\mathbf{i}$ belonging to class $\mathbf{j}$,
\begin{equation}
    \mathbf{Y^\prime}=\operatorname{softmax}(\mathbf{B} \cdot \mathbf{R} + \mathbf{a}).
\end{equation}
Suppose there are $\mathcal{T}$ nodes with labels in the training set. We adopt cross-entropy to measure 
the difference between prediction label $\mathbf{Y^\prime_{ij}}$ and ground truth label $\mathbf{Y_{ij}}$, i.e.,
\begin{equation}
    {L}_{cl}=-\sum_{\mathbf{i=1}}^{\mathcal{T}} \sum_{\mathbf{j=1}}^{\mathbf{M}} \mathbf{Y_{i j}} \ln \mathbf{Y^\prime_{i j}}
    \label{equ: ce_loss}.
\end{equation}

Finally, by combining $L_{cl}$, $L_{re}$ and $L_{cr}$, the overall loss function of our SPGRL model can be represented as:
\begin{equation}
    L = L_{cl} + \alpha {L}_{re} + \beta L_{cr},
\label{equ: all_loss}
\end{equation}
where $\alpha$ and $\beta$ are trade-off hyper-parameters. The parameters of the whole framework are updated via backpropagation.
The detailed description of our algorithm is provided in 
Algorithm \ref{algo: algo}.

\begin{algorithm}
    \caption{The proposed algorithm SPGRL}
    \label{algo: algo}
    \SetKwFunction{isOddNumber}{isOddNumber}
    \SetKwInput{Input}{Input}
    \SetKwInput{Output}{Output}

    \KwIn{Node feature matrix $\mathbf{X}$; original graph adjacency matrix $\mathbf{A}$; 
    node label matrix $\mathbf{Y}$; maximum number of iterations $\eta$}
    
    Compute the feature graph topological structure $\mathbf{\hat{A}}$ according to $\mathbf{X}$ by running $k$NN algorithm.
    
    \For{$it=1$ \KwTo $\eta$}{
    
    $\mathbf{Z^t}$ = $GCN$($\mathbf{A},\mathbf{X}$) 
    
    $\mathbf{Z^f}$ = $GCN^{\prime}$($\mathbf{\hat{A}},\mathbf{X}$) \tcp{embeddings of two graphs}
   
    $\mathbf{Z^t}$ and $\mathbf{Z^f}$ interact with local node-level information.  
    
    $q_{\phi}(\mathbf{\hat{A}}|\mathbf{Z^t})$ = $Decoder$($\mathbf{Z^t}$)
    
    $q_{\phi}(\mathbf{A}|\mathbf{Z^f})$ = $Decoder^{\prime}$($\mathbf{Z^f}$)
    \tcp{reconstructing two graphs}
   
    $q_{\phi}(\mathbf{\hat{A}}|\mathbf{Z^t})$ constrained by $\mathbf{\hat {A}}$, $q_{\phi}(\mathbf{A}|\mathbf{Z^f})$ constrained by $\mathbf{A}$
    
    
     


    
    
    Calculate the overall loss with Eq.(\ref{equ: all_loss})
    
    Update all parameters of framework according to
the overall loss
    }
    
    Predict the labels of unlabeled nodes based on the trained framework.
    
    \KwOut{Classification result $\mathbf{Y^{\prime}}$}
\end{algorithm}

\begin{figure*}[!t]
\centering
\subfigure[ACM Epoch 0]{
		\includegraphics[width=0.14\textwidth]{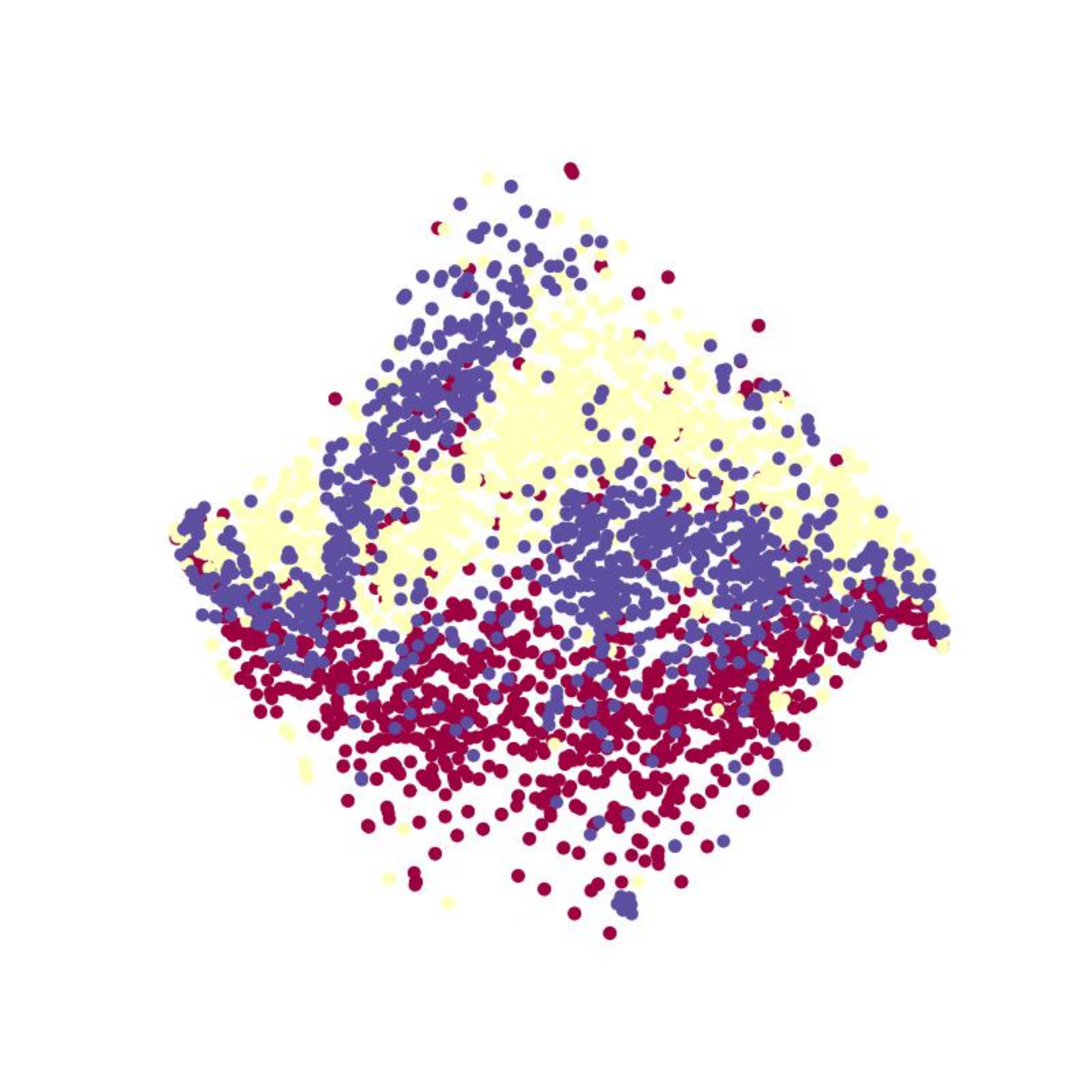}
	}
\subfigure[ACM Epoch 30]{
		\includegraphics[width=0.14\textwidth]{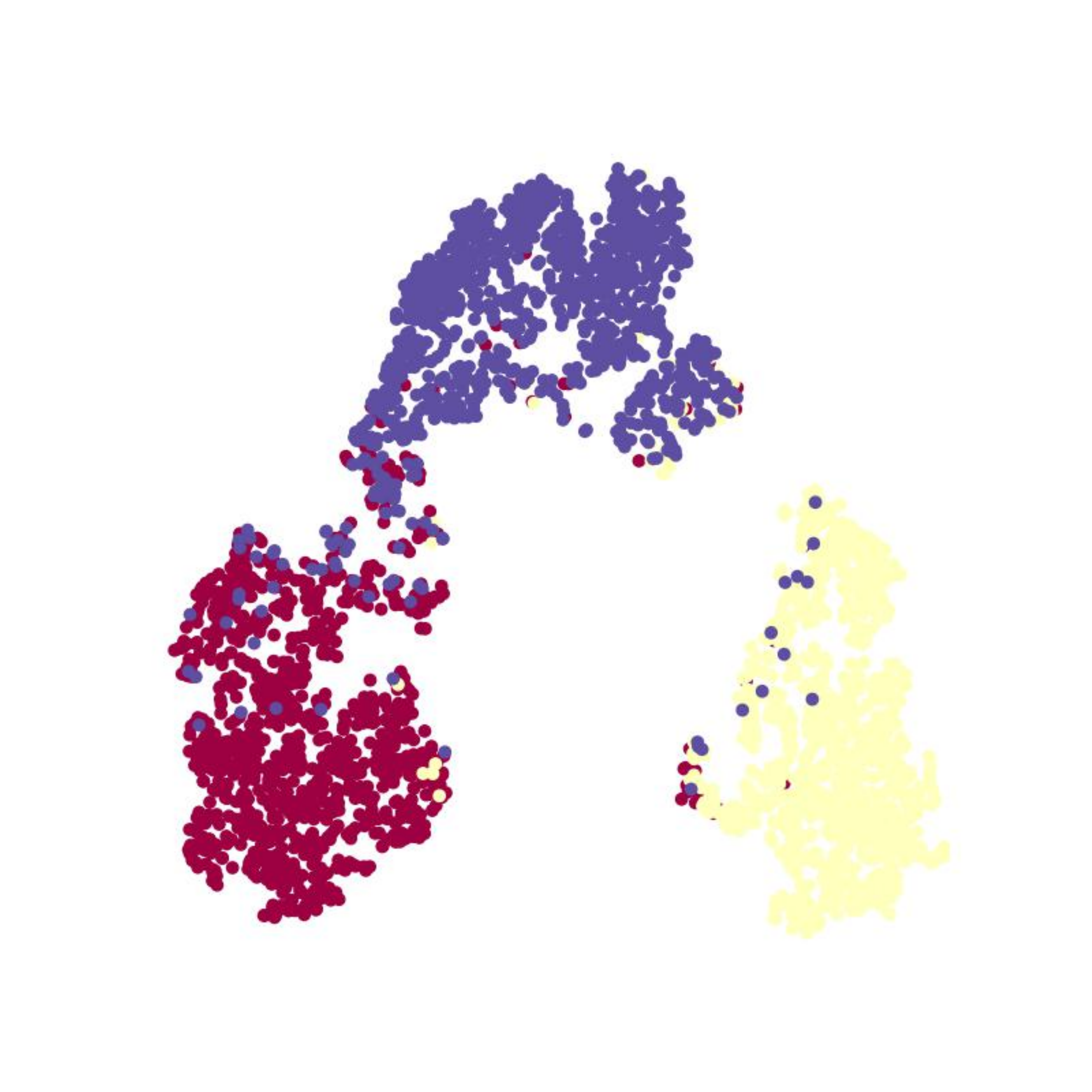}
	}
\subfigure[BlogCatalog Epoch 0]{
		\includegraphics[width=0.15\textwidth]{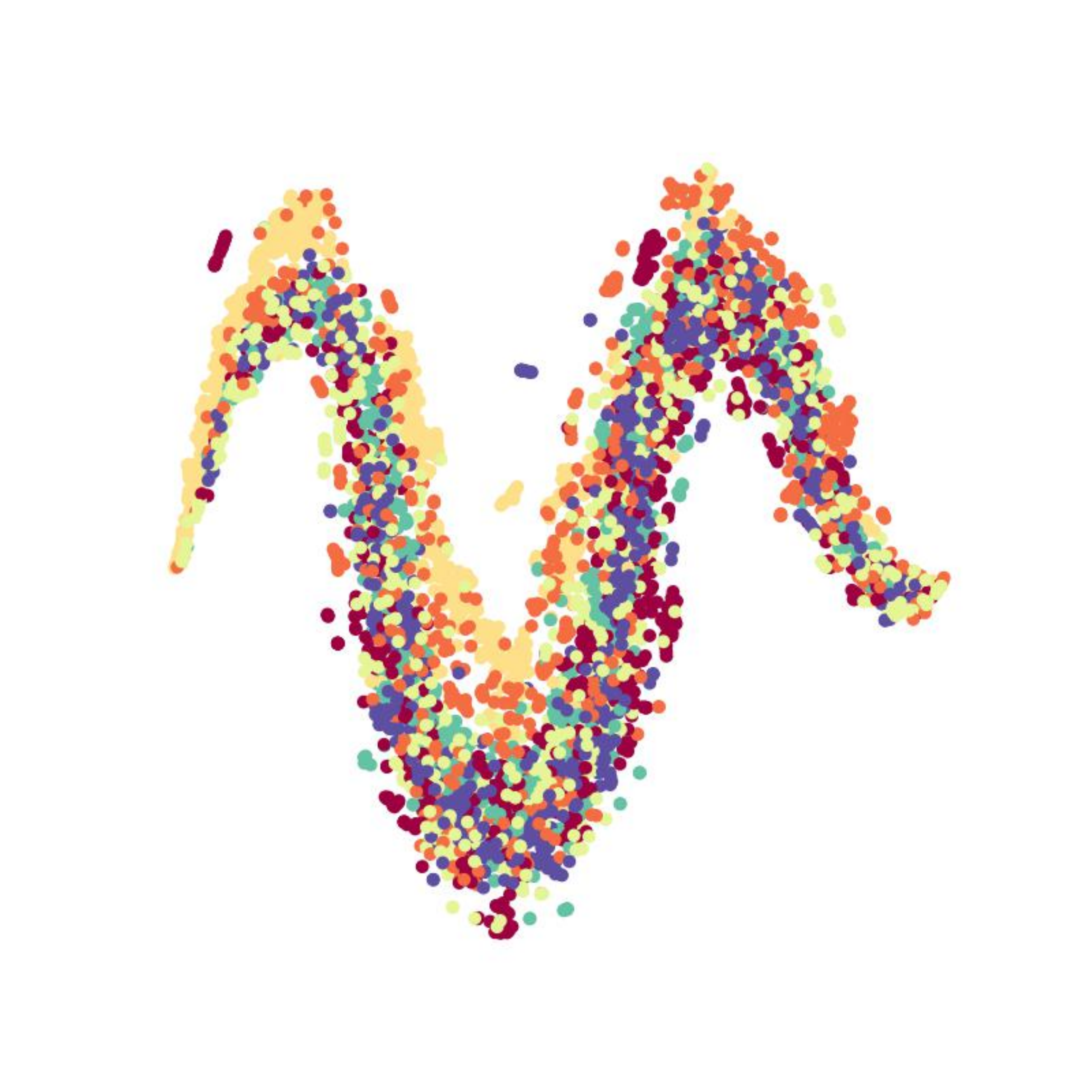}
	}
\subfigure[BlogCatalog Epoch 30]{
		\includegraphics[width=0.16\textwidth]{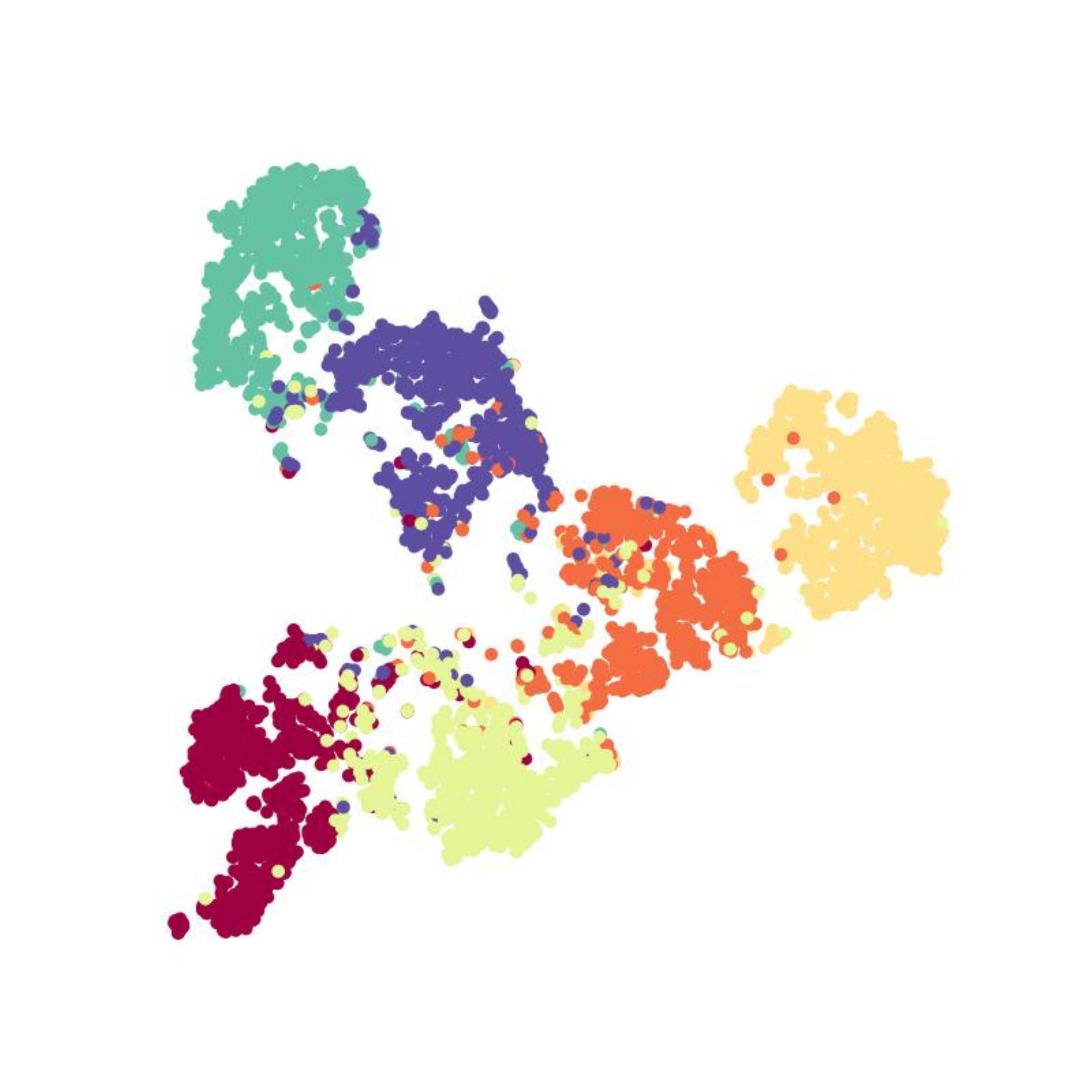}
	}
\subfigure[Citeseer Epoch 0]{
		\includegraphics[width=0.14\textwidth]{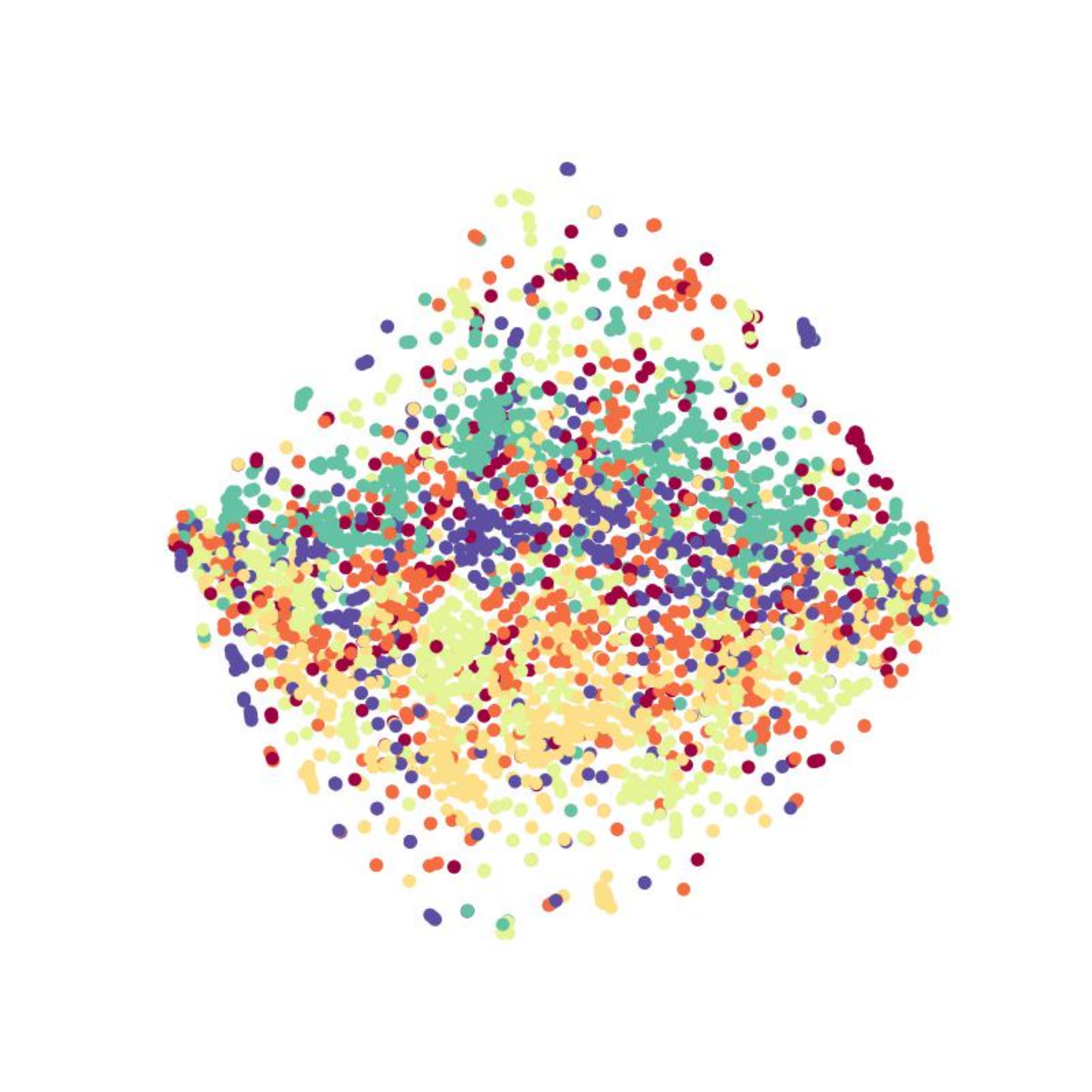}
	}
\subfigure[Citeseer Epoch 30]{
		\includegraphics[width=0.14\textwidth]{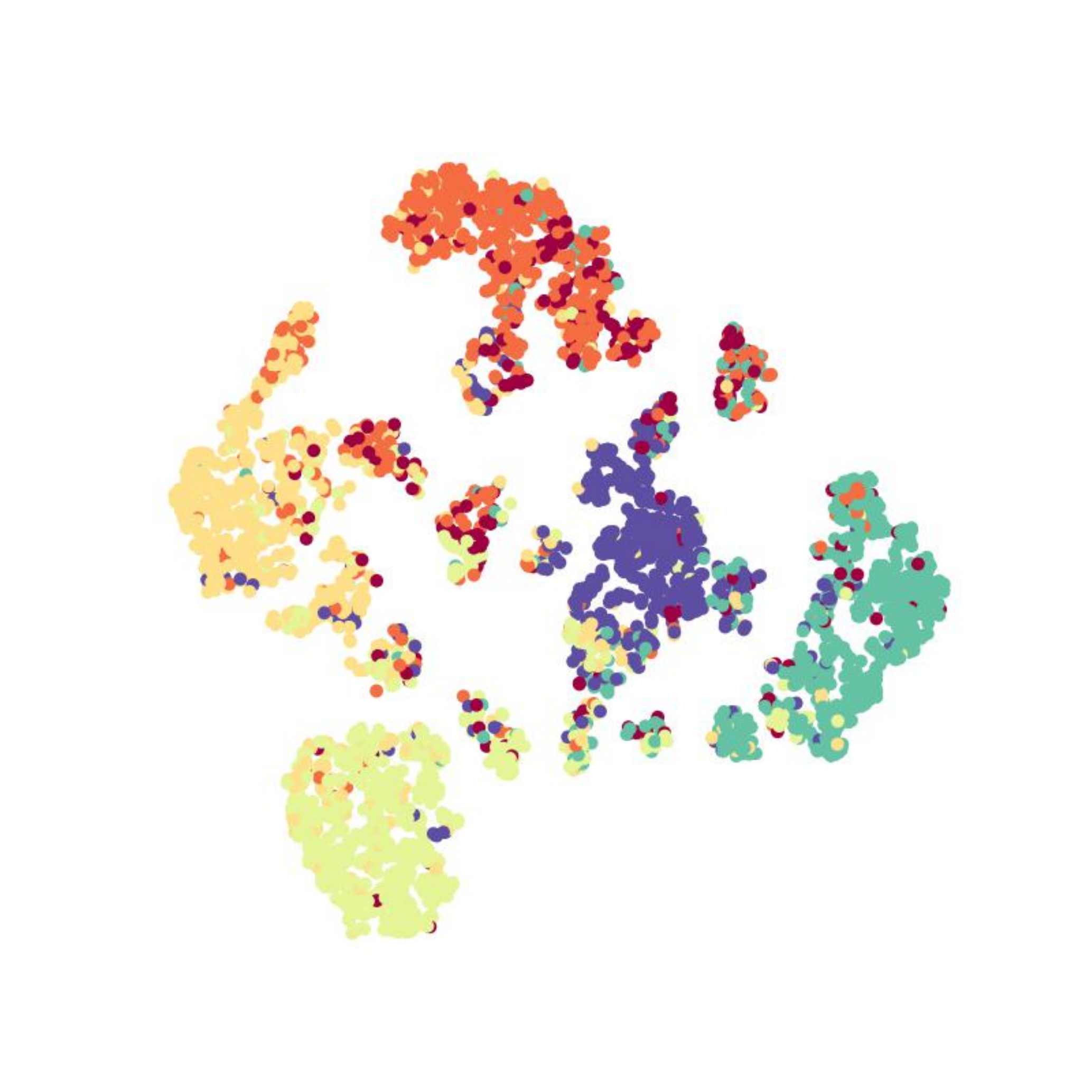}
	}
\caption{The t-SNE visualisation of node representations of ACM, BlogCatalog, and Citeseer during training.}
\label{fig:tsne graph during training}
\end{figure*}

\begin{table}[H]
\renewcommand{\arraystretch}{1.0}
\caption{The statistics of datasets.}
\label{table:dataset}
 \resizebox{\linewidth}{!}{
\begin{tabular}{ccccc}
\hline
\textbf{Datasets}    & \textbf{Nodes} & \textbf{Edges} & \textbf{Dimensions} & \textbf{Classes} \\ \hline
\textbf{Citeseer}    & 3327           & 4732           & 3703                & 6                \\
\textbf{PubMed}      & 19717          & 44338          & 500                 & 3                \\
\textbf{ACM}         & 3025           & 13128          & 1870                & 3                \\
\textbf{BlogCatalog} & 5196           & 171743         & 8189                & 6                \\
\textbf{UAI2010}     & 3067           & 28311          & 4973                & 19               \\
\textbf{Flickr}      & 7575           & 239738         & 12047               & 9                \\ \hline
\end{tabular}}
\end{table}

\begin{figure*}[!htbp]
\centering
\subfigure[GCN]{
		\includegraphics[width=0.18\textwidth]{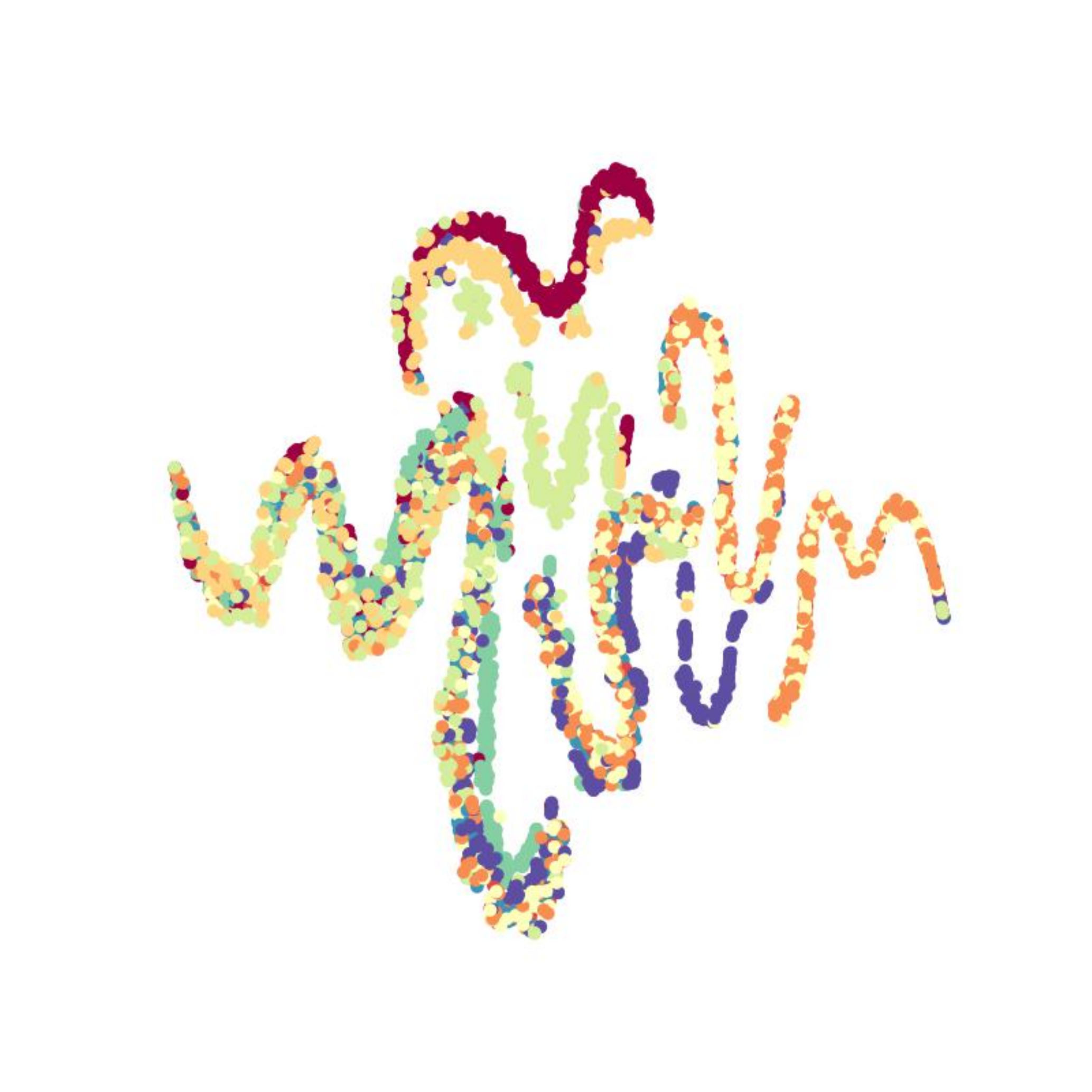}
	}
\subfigure[AMGCN]{
		\includegraphics[width=0.18\textwidth]{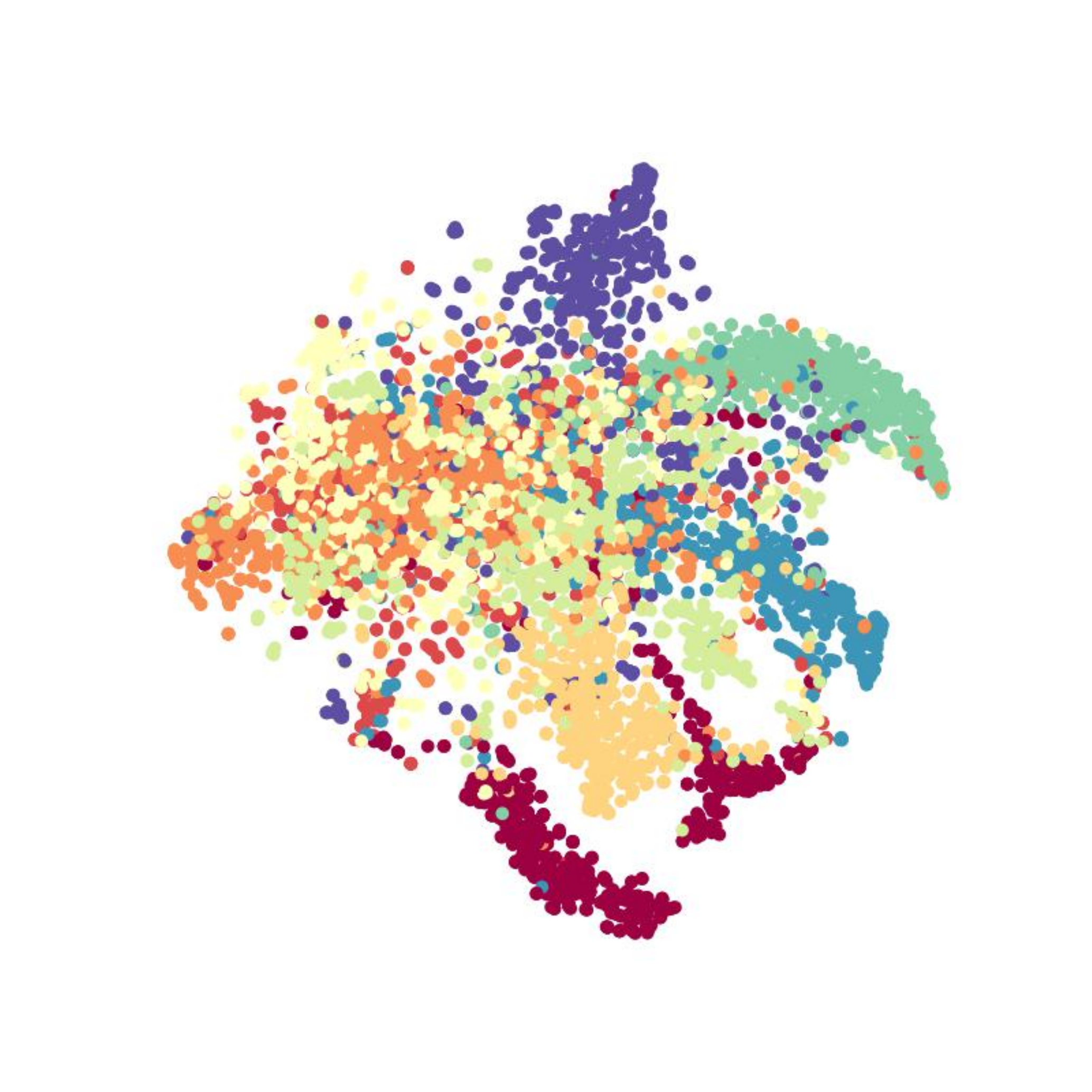}
	}
\subfigure[SLAPS]{
		\includegraphics[width=0.18\textwidth]{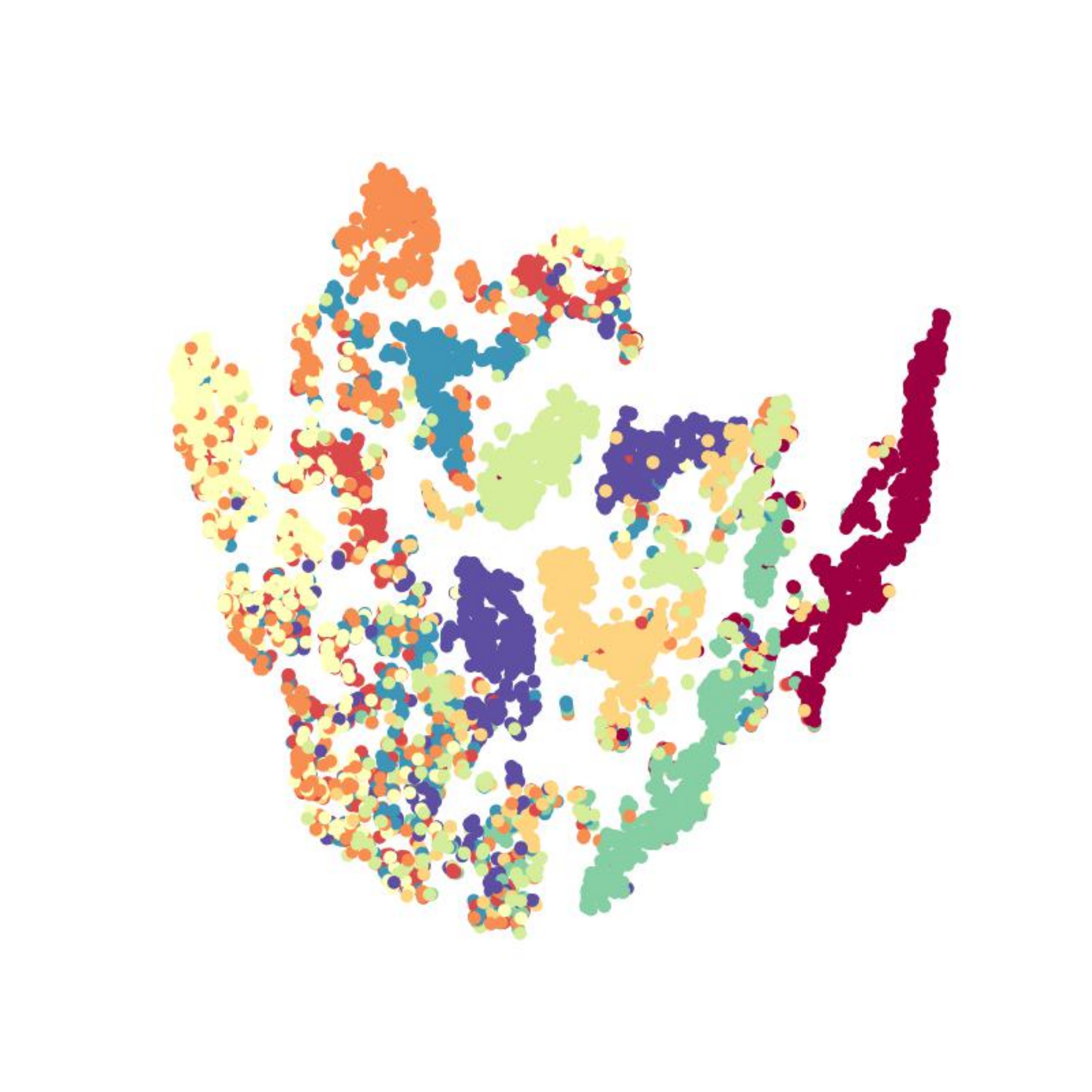}
	}
\subfigure[SCRL]{
		\includegraphics[width=0.18\textwidth]{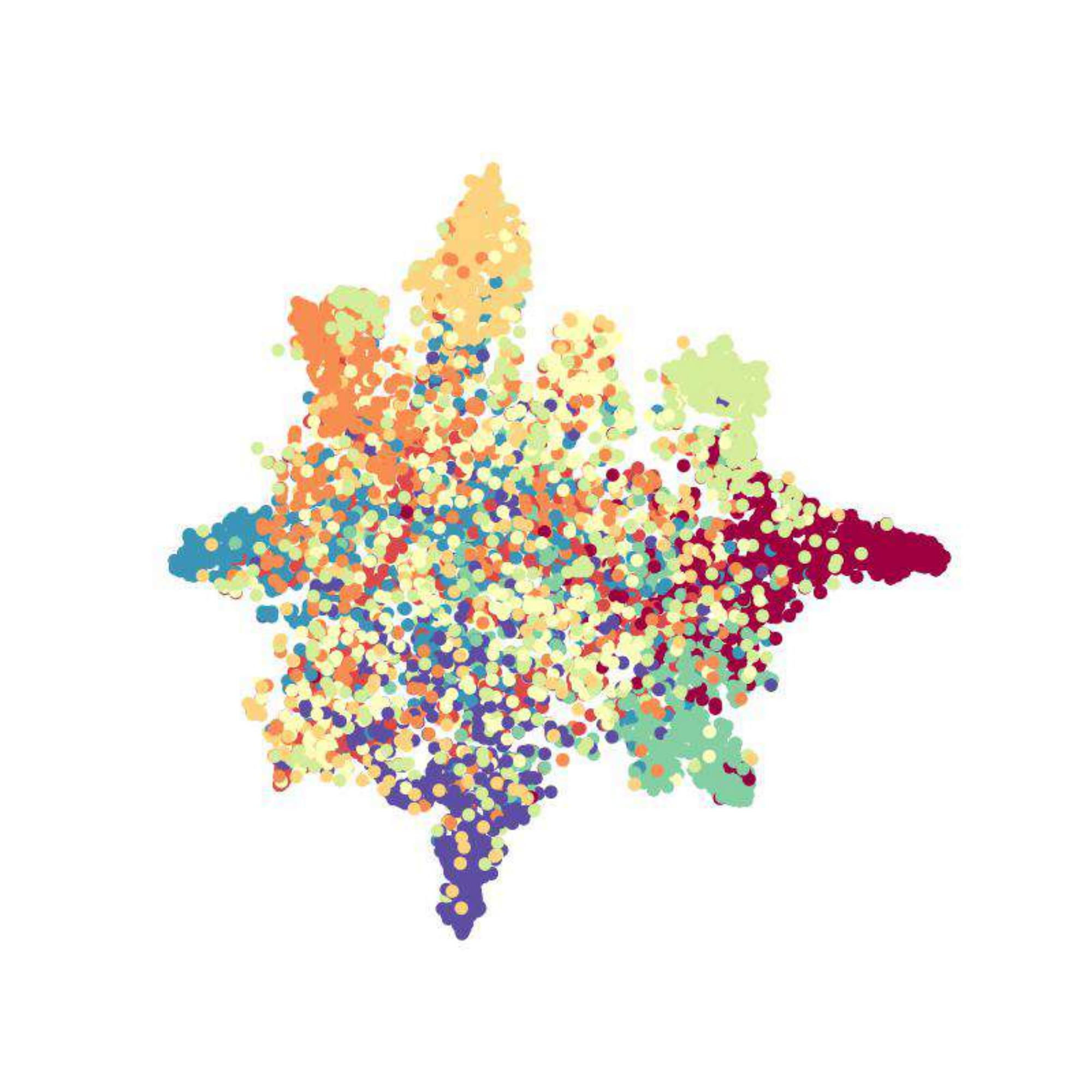}
	}
\subfigure[SPGRL]{
		\includegraphics[width=0.18\textwidth]{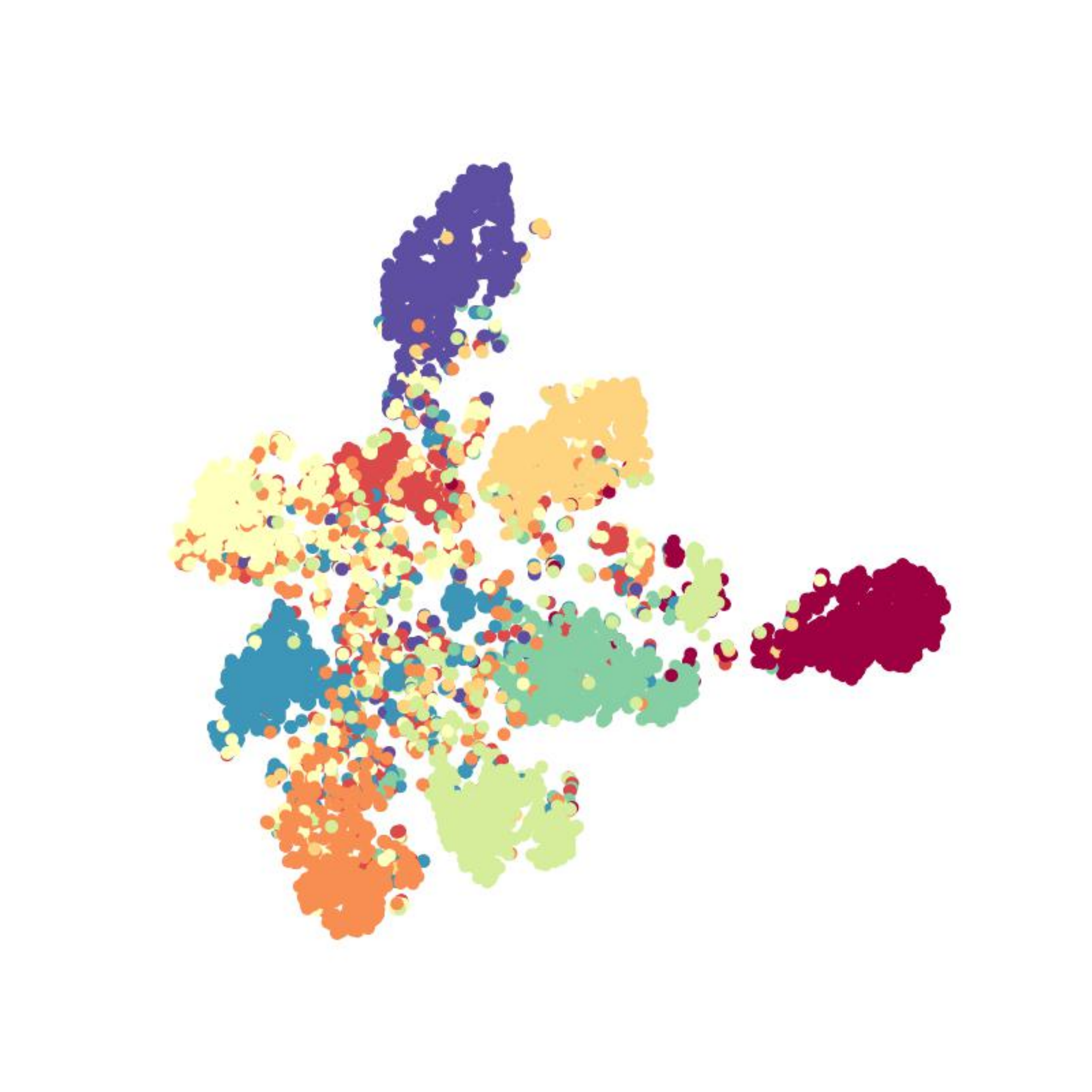}
	}
\caption{Visualization of learnt representations of different methods on Flickr dataset.}
\label{fig:tsne graph with different methods}
\end{figure*}

\begin{table*}[!t]
\caption{Node classification results($\%$). L/C refers to the number of labeled nodes per class. }
\label{table:result}
\centering
\renewcommand{\arraystretch}{1.0}
\begin{tabular}{|c|cc|cc|cc|cc|cc|cc|}
\hline
Dataset       & \multicolumn{6}{c|}{\textbf{ACM}}                                                                                                 & \multicolumn{6}{c|}{\textbf{BlogCatalog}}                                                                                                              \\ \hline
L/C           & \multicolumn{2}{c|}{20}                   & \multicolumn{2}{c|}{40}                   & \multicolumn{2}{c|}{60}                   & \multicolumn{2}{c|}{20}                          & \multicolumn{2}{c|}{40}                          & \multicolumn{2}{c|}{60}                          \\ \hline
Label Rate           & \multicolumn{2}{c|}{1.98\%}                   & \multicolumn{2}{c|}{3.97\%}                   & \multicolumn{2}{c|}{5.95\%}                   & \multicolumn{2}{c|}{2.31\%}                          & \multicolumn{2}{c|}{4.62\%}                          & \multicolumn{2}{c|}{6.93\%}                          \\ \hline
Metrics       & \multicolumn{1}{c|}{ACC} & F1             & \multicolumn{1}{c|}{ACC} & F1             & \multicolumn{1}{c|}{ACC} & F1             & \multicolumn{1}{c|}{ACC} & F1                    & \multicolumn{1}{c|}{ACC} & F1                    & \multicolumn{1}{c|}{ACC} & F1                    \\ \hline
DeepWalk~\cite{perozzi2014deepwalk}     & 62.69     & 62.11       & 63.00      & 61.88     & 67.03        & 66.99         & 38.67           & 34.96          & 50.80          & 48.61         & 55.02           & 53.36   \\
LINE~\cite{tang2015line}                & 41.28     & 40.12       & 45.83      & 45.79     & 50.41        & 49.92         & 58.75           & 57.75          & 61.12          & 60.72         & 64.53           & 63.81   \\
ChebNet~\cite{defferrard2016convolutional} & 75.24  & 74.86       & 81.64      & 81.26     & 85.43        & 85.26         & 38.08           & 33.39          & 56.28          & 53.86         & 70.06           & 68.37   \\
GCN~\cite{kipf2017semi}                 & 87.80     & 87.82       & 89.06      & 89.00     & 90.54        & 90.49         & 69.84           & 68.73          & 71.28          & 70.71         & 72.66           & 71.80   \\
$k$NN-GCN~\cite{wang2020gcn}              & 78.52     & 78.14       & 81.66      & 81.53     & 82.00        & 81.95         & 75.49           & 72.53          & 80.84          & 80.16         & 82.46           & 81.90   \\
GAT~\cite{velickovic2018graph}         & 87.36     & 87.44       & 88.60      & 88.55     & 90.40        & 90.39         & 64.08           & 63.38          & 67.40          & 66.39         & 69.95           & 69.08   \\
Demo-Net~\cite{wu2019net}               & 84.48     & 84.16       & 85.70      & 84.83     & 86.55        & 84.05         & 54.19           & 52.79          & 63.47          & 63.09         & 76.81           & 76.73   \\
MixHop~\cite{abu2019mixhop}             & 81.08     & 81.40       & 82.34      & 81.13     & 83.09        & 82.24         & 65.46           & 64.89          & 71.66          & 70.84         & 77.44           & 76.38   \\
DGI~\cite{veli2018deep}           &90.48	    &90.40	      &90.97   	   &90.88	   &90.94	      &90.79	      &64.59	        &63.58	         &65.09	          &64.15	      &65.90	        &65.00    \\
GRACE~\cite{Zhu:2020vf}                  & 89.04     & 89.00       & 89.46      & 89.36     & 91.08        & 91.03         & 76.56           & 75.56          & 76.66          & 75.88         & 77.66           & 77.08   \\
AMGCN~\cite{wang2020gcn}                & 90.40     & 90.43       & 90.76      & 90.66     & 91.42        & 91.36         & 81.89           & 81.36          & 84.94          & 84.32         & 87.30           & 86.94   \\
GMI~\cite{peng2020graph}                &90.22	     &90.00	    &90.68	       &90.64	   &91.48	      &91.45           &66.46	&39.2	                 &68.01         &40.42	          &72.59	 &43.24      \\  
SCRL~\cite{liu2021self}                 & 91.82     & 91.79       & 92.06      & 92.04     & 92.82        & 92.80         & 90.22           & 89.89          & 90.26          & 89.90         & 91.58           & 90.76         \\
SLAPS ~\cite{fatemi2021slaps}           & 65.32	    & 60.00	      & 55.46	   &47.73	    &60.13	      &52.56          &87.80	      &87.34         	&88.50	          &87.57	      &89.50	          &89.22   \\ 
GCA~\cite{zhu2021graph}                 &88.39	&8879	&91.95	 &90.99	&91.75	&90.79    &80.51	&81.28	&84.89	&84.04	&86.34	&86.19
\\ \hline
\textbf{SPGRL} & \textbf{93.30}         & \textbf{93.27} & \textbf{93.50}     & \textbf{93.48}  & \textbf{94.00}  & \textbf{93.98} & \textbf{90.70}  & \textbf{90.12}   & \textbf{92.10}  & \textbf{91.34}  & \textbf{92.30}& \textbf{92.13}        \\ \hline
Dataset       & \multicolumn{6}{c|}{\textbf{Flickr}}      & \multicolumn{6}{c|}{\textbf{UAI2010}}           \\ \hline
L/C           & \multicolumn{2}{c|}{20}   & \multicolumn{2}{c|}{40}   & \multicolumn{2}{c|}{60}    & \multicolumn{2}{c|}{20}    & \multicolumn{2}{c|}{40}        & \multicolumn{2}{c|}{60}       \\ \hline
Label Rate    & \multicolumn{2}{c|}{2.38\%}  & \multicolumn{2}{c|}{4.75\%}      & \multicolumn{2}{c|}{7.13\%}   & \multicolumn{2}{c|}{12.39\%}  & \multicolumn{2}{c|}{24.78\%}  & \multicolumn{2}{c|}{37.17\%}       \\ \hline
Metrics       & \multicolumn{1}{c|}{ACC} & F1    & \multicolumn{1}{c|}{ACC} & F1    & \multicolumn{1}{c|}{ACC} & F1    & \multicolumn{1}{c|}{ACC} & F1  & \multicolumn{1}{c|}{ACC} & F1    & \multicolumn{1}{c|}{ACC} & F1                    \\ \hline
DeepWalk~\cite{perozzi2014deepwalk}     & 24.33     & 21.33      & 28.79       & 26.90     & 30.10        & 27.28         & 42.02           & 32.92           & 51.26         & 46.01         & 54.37           & 44.43   \\
LINE~\cite{tang2015line}                & 33.25     & 31.19      & 37.67       & 37.12     & 38.54        & 37.77         & 43.47           & 37.01           & 45.37         & 39.62         & 51.05           & 43.76   \\
ChebNet~\cite{defferrard2016convolutional} & 23.26  & 21.27      & 35.10       & 33.53     & 41.70        & 40.17         & 50.02           & 33.65           & 58.18         & 38.80         & 59.82           & 40.60   \\
GCN~\cite{kipf2017semi}                 & 41.42     & 39.95      & 45.48       & 43.27     & 47.96        & 46.58         & 49.88           & 32.86           & 51.80         & 33.80         & 54.40           & 32.14   \\
$k$NN-GCN~\cite{wang2020gcn}              & 69.28     & 70.33      & 75.08       & 75.40     & 77.94        & 77.97         & 66.06           & 52.43           & 68.74         & 54.45         & 71.64           & 54.78   \\
GAT~\cite{velickovic2018graph}         & 38.52     & 37.00      & 38.44       & 36.94     & 38.96        & 37.35         & 56.92           & 39.61           & 63.74         & 45.08         & 68.44           & 48.97   \\
Demo-Net~\cite{wu2019net}               & 34.89     & 33.53      & 46.57       & 45.23     & 57.30        & 56.49         & 23.45           & 16.82           & 30.29         & 26.36         & 34.11           & 29.03   \\
MixHop~\cite{abu2019mixhop}             & 39.56     & 40.13      & 55.19       & 56.25     & 64.96        & 65.73         & 61.56           & 49.19           & 65.05         & 53.86         & 67.66           & 56.31   \\
DGI~\cite{veli2018deep}           &34.95   	&33.1     	&34.98	       &33.07	   &35.51		  &34.37		  &33.26		    &11.86		      &32.55		  &9.29		      &32.44		    &9.37    \\
GRACE~\cite{Zhu:2020vf}                  & 49.42     & 48.18      & 53.64       & 52.61     & 55.67        & 54.61         & 65.54           & 48.38           & 66.67         & 49.50         & 68.68           & 51.51   \\
AMGCN~\cite{wang2020gcn}                & 75.26     & 74.63      & 80.06       & 79.36     & 82.10        & 81.81         & 70.10           & 55.61           & 73.14         & 64.88         & 74.40           & 65.99   \\ 
GMI~\cite{peng2020graph}                &49.17	   &28.43	     &52.74	       &30.94	    &53.78	      &31.50           &60.69	         &46.75	          &63.14	       &49.10	       &64.73	       &44.36\\
SCRL~\cite{liu2021self}               & 79.52     & 78.89      & 84.23       & 84.03     & 84.54        & 84.51         & 72.90           & 57.80           & 74.58         & 67.40         & 74.90           & 67.54\\
SLAPS ~\cite{fatemi2021slaps}           & 72.20	    & 72.48	      & 79.00	   &78.90	    &76.20	      &76.50          &46.82	      &41.60         	&34.62	          &25.28	      &62.51	          &51.81   \\ 
GCA~\cite{zhu2021graph}                    &63.44	&63.26 	&63.90	&64.60	&64.43	&64.64                                            &72.55	&56.97	&73.27	 &54.55  	&73.60	&56.00\\ \hline
\textbf{SPGRL} & \textbf{82.20}      & \textbf{81.24} & \textbf{86.20}   & \textbf{85.93} & \textbf{87.10}   & \textbf{85.97} & \textbf{76.30}   & \textbf{61.49}   & \textbf{78.20}  & \textbf{68.73}    & \textbf{79.80}      & \textbf{71.38}        \\ \hline
Dataset       & \multicolumn{6}{c|}{\textbf{Citeseer}}                & \multicolumn{6}{c|}{\textbf{PubMed}}                                            \\ \hline
L/C           & \multicolumn{2}{c|}{20}      & \multicolumn{2}{c|}{40}     & \multicolumn{2}{c|}{60}      & \multicolumn{2}{c|}{20}   & \multicolumn{2}{c|}{40}     & \multicolumn{2}{c|}{60}          \\ \hline
Label Rate    & \multicolumn{2}{c|}{3.61\%}      & \multicolumn{2}{c|}{7.21\%}    & \multicolumn{2}{c|}{10.82\%}       & \multicolumn{2}{c|}{0.30\%}     & \multicolumn{2}{c|}{0.61\%}      & \multicolumn{2}{c|}{0.91\%}                          \\ \hline
Metrics       & \multicolumn{1}{c|}{ACC} & F1      & \multicolumn{1}{c|}{ACC} & F1         & \multicolumn{1}{c|}{ACC} & F1         & \multicolumn{1}{c|}{ACC} & F1       & \multicolumn{1}{c|}{ACC} & F1            & \multicolumn{1}{c|}{ACC} & F1                    \\ \hline
DeepWalk~\cite{perozzi2014deepwalk}     & 43.47     & 38.09      & 45.15       & 43.18     & 48.86        & 48.01          & -              & -               & -             & -            & -                & -       \\
LINE~\cite{tang2015line}                & 32.71     & 31.75      & 33.32       & 32.42     & 35.39        & 34.37          & -              & -               & -             & -            & -                & -       \\
ChebNet~\cite{defferrard2016convolutional}  & 69.80 & 65.92      & 71.64       & 68.31     & 73.26        & 70.31          & 74.20          & 73.51           & 76.00         & 74.92        & 76.51            & 75.83   \\
GCN~\cite{kipf2017semi}                 & 70.30     & 67.50      & 73.10       & 69.70     & 74.48        & 71.24          & 79.00          & 78.45           & 79.98         & 79.17        & 80.06            & 79.65   \\
$k$NN-GCN~\cite{wang2020gcn}              & 61.35     & 58.86      & 61.54       & 59.33     & 62.38        & 60.07          & 71.62          & 71.92           & 74.02         & 74.09        & 74.66            & 75.18   \\
GAT~\cite{velickovic2018graph}         & 72.50     & 68.14      & 73.04       & 69.58     & 74.76        & 71.60          & -              & -               & -             & -            & -                & -       \\
Demo-Net~\cite{wu2019net}               & 69.50     & 67.84      & 70.44       & 66.97     & 71.86        & 68.22          & -              & -               & -             & -            & -                & -       \\
MixHop~\cite{abu2019mixhop}             & 71.40     & 66.96      & 71.48       & 67.40     & 72.16        & 69.31          & -              & -               & -             & -            & -                & -       \\
DGI~\cite{veli2018deep}           &71.24   	&67.05     	&71.26	       &67.75	   &73.92		  &70.26	       & -              & -               & -             & -            & -                & -       \\
GRACE~\cite{Zhu:2020vf}                  & 71.70     & 68.14      & 72.38       & 68.74     & 74.20        & 70.73          & 79.50          & 79.33  & 80.32         & 79.64        &80.24             & 80.33   \\
AMGCN~\cite{wang2020gcn}                & 73.10     & 68.42      & 74.70       & 69.81     & 75.56        & 70.92          & 76.18          & 76.86           & 77.14         & 77.04        & 77.74            & 77.09   \\ 
GMI~\cite{peng2020graph}                &71.24	     &67.1	       &73.1	    &68.57	   &73.96	       &70.25      & -              & -               & -             & -            & -                & -       \\
SCRL~\cite{liu2021self}                             & 73.62     & 69.78      & 75.08       & 70.68     & 75.96        & 72.84          & 79.62          & 78.88           & 80.74         & 80.24        & 81.03            & 80.55 \\ 
SLAPS ~\cite{fatemi2021slaps}           & 70.50	    & 67.23	      & 72.10	   &69.15	    &73.00	      &69.80          &71.70	      &72.29         	&71.60	          &71.56	      &70.60	          &71.16   \\ 
GCA~\cite{zhu2021graph}                     &71.39	&68.46	&72.96	 &68.02  	&73.92	&69.10   &\textbf{82.00}	 &\textbf{81.50}	   &\textbf{82.59}	&\textbf{82.43}	&82.03	&81.75                               \\ \hline
\textbf{SPGRL}    & \textbf{75.90}   & \textbf{70.98}  & \textbf{77.40}     & \textbf{73.75}    & \textbf{78.30}   & \textbf{73.98}  & 77.60   & 76.98    & 81.20   & 81.01   & \textbf{82.10}           & \textbf{81.94}        \\ \hline
\end{tabular}
\end{table*}

\section{EXPERIMENT}
In this section, we conduct extensive experiments to evaluate the effectiveness of the proposed method.
\subsection{Datasets}
We select four commonly used citation networks (UAI2010~\cite{wang2018unified}, ACM~\cite{wang2019heterogeneous}, Citeseer~\cite{kipf2017semi}, PubMed 
~\cite{sen2008collective}) and two social networks (BlogCatalog~\cite{meng2019co}, Flickr~\cite{meng2019co}). 
Specifically, Citeseer consists of 3327 scientific publications extracted from the Citeseer digital library classified into one of six classes. 
PubMed consists of 19717 scientific publications from PubMed database pertaining to diabetes classified into one of three classes. 
ACM network is extracted from ACM database where publications are represented by nodes and those with the same author are connected by edges. 
UAI2010 contains 3067 nodes in 19 classes. 
BlogCatalog is a social blog directory containing links between 5196 blogs. 
Flickr is  widely used by photo researchers and bloggers to host images that they embed in blogs and social media. Flickr is composed of 7575 users and they are classified into nine groups. 
The statistical information of datasets is summarized in Table \ref{table:dataset}.
\subsection{Experimental Setup}
The experiments are implemented in the PyTorch platform using an Intel(R) Xeon(R) Gold 5218 CPU, 
and GeForce RTX 3090 24G GPU.
Technically, two layers GCN is built and we train
our model by utilizing the Adam~\cite{adam} optimizer with learning rate ranging
from 0.0001 to 0.0005. In order to prevent over-fitting, we set the dropout rate to 0.5.
In addition, we set weight decay $\in \left\{1e-4, \cdots, 5e-3 \right\}$ and $k$ $\in \left\{2, \cdots, 20 \right\}$ for $k$NN graph. For fairness, we follow Wang \emph{et al.}~\cite{wang2020gcn} and 
select 20, 40, 60 nodes per class for training and 1000 nodes for testing.
For example, there are 6 types of nodes in Citeseer, therefore we train our model on training set with 120/240/360 nodes, corresponding to label rate of 3.61\%, 7.21\%, 10.82\%, respectively. Two popular metrics are applied to quantitatively evaluate the semi-supervised node classification: Accuracy (ACC) and F1-Score (F1). 
We repeatedly train and test our model for five times with the same partition of dataset and then report the average of ACC and F1.


\subsection{Baselines}
We choose some representative methods to compare.
\begin{itemize}
    \item \textbf{DeepWalk}~~\cite{perozzi2014deepwalk} is a graph embedding method that 
    merely takes into account the structure of the graph.
    \item \textbf{LINE}~ ~\cite{tang2015line} is an efficient large-scale network embedding method preserving first-order and second-order proximity of the network separately. 
    \item \textbf{ChebNet}~ ~\cite{defferrard2016convolutional} is a spectral-based GCN that uses Chebyshev polynomial to reduce computational complexity.
    \item \textbf{GCN}~ ~\cite{kipf2017semi} further solves the efficiency problem by 
    introducing first-order approximation of ChebNet. 
    \item \textbf{$k$NN-GCN}~ ~\cite{wang2020gcn} use the sparse $k$-nearest neighbor graph 
    calculated from feature matrix as the input graph of GCN and name it $k$NN-GCN.
    \item \textbf{GAT}~ ~\cite{velickovic2018graph} adopts attention mechanism to learn the 
    relative weights between two connected nodes.
   \item \textbf{Demo-Net}~~\cite{wu2019net} is a degree-specific graph neural network
    for node classification.
    \item \textbf{MixHop}~ ~\cite{abu2019mixhop} is a GCN-based method that concatenates embeddings aggregated
    using the transition matrices of $k$-hop random walks before each layer.
    \item \textbf{DGI}~ ~\cite{veli2018deep} makes local information maximization representation between spanning graphs.
    \item \textbf{GRACE}~ ~\cite{Zhu:2020vf}  is a graph contrastive representation learning framework, which maximizes the MI of two graph representation at the node level by constructing a pair of corrupted graph.
    \item \textbf{AMGCN}~ ~\cite{wang2020gcn} extracts embeddings from node feature value, topological structure, and uses the attention mechanism to learn the adaptive  weights of embeddings.
     \item \textbf{GMI}~ ~\cite{peng2020graph} maximizes the MI between the original graph and output graph from the perspective of node and edge.
    \item \textbf{SCRL}~ ~\cite{liu2021self} proposes a self-supervised framework to learn a consensus representation for attributed graph to exploit the topology structure and feature information of the graph.
    \item \textbf{SLAPS}~ ~\cite{fatemi2021slaps}  defines a collective homogeneous node graph structure and regularizes it to guide the learning model to solve the supervision starvation problem.
    \item \textbf{GCA}~~\cite{zhu2021graph} proposes an adaptive data augmentation scheme to preserve the intrinsic structure and properties of the graph by exploiting the connection patterns of original graph. 
\end{itemize}

\begin{figure}[!htbp]
\centering
  \includegraphics[width=0.9\linewidth]{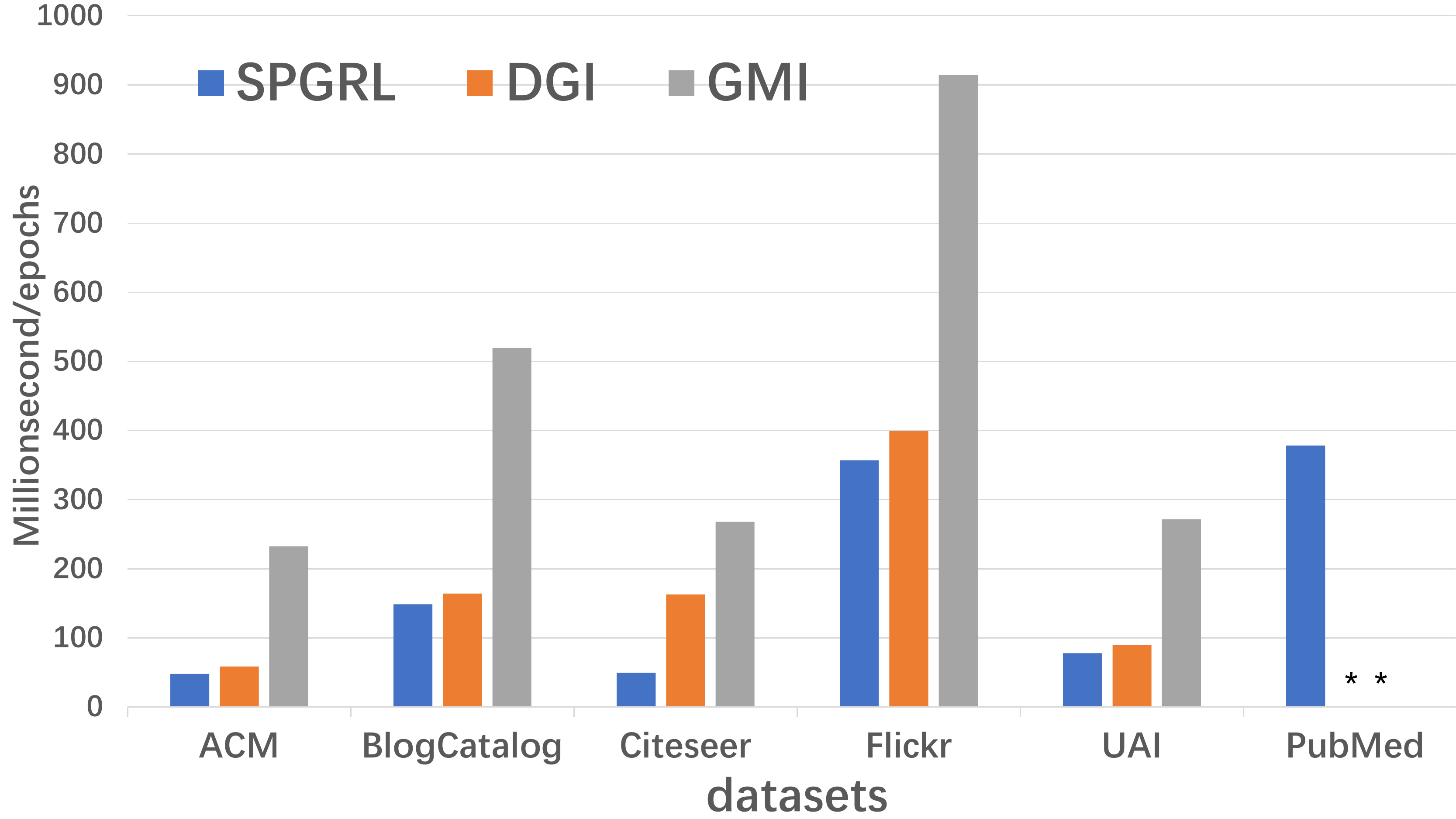}
  \caption{Averaged time cost per epoch of SPGRL, DGI and GMI for six datasets. 
  (*) indicates out-of-memory error and vertical axis is in log-scale.}
  \label{fig: time}
\end{figure}

\begin{figure*}[!htbp]
    \centering
    \includegraphics[width=0.9\textwidth]{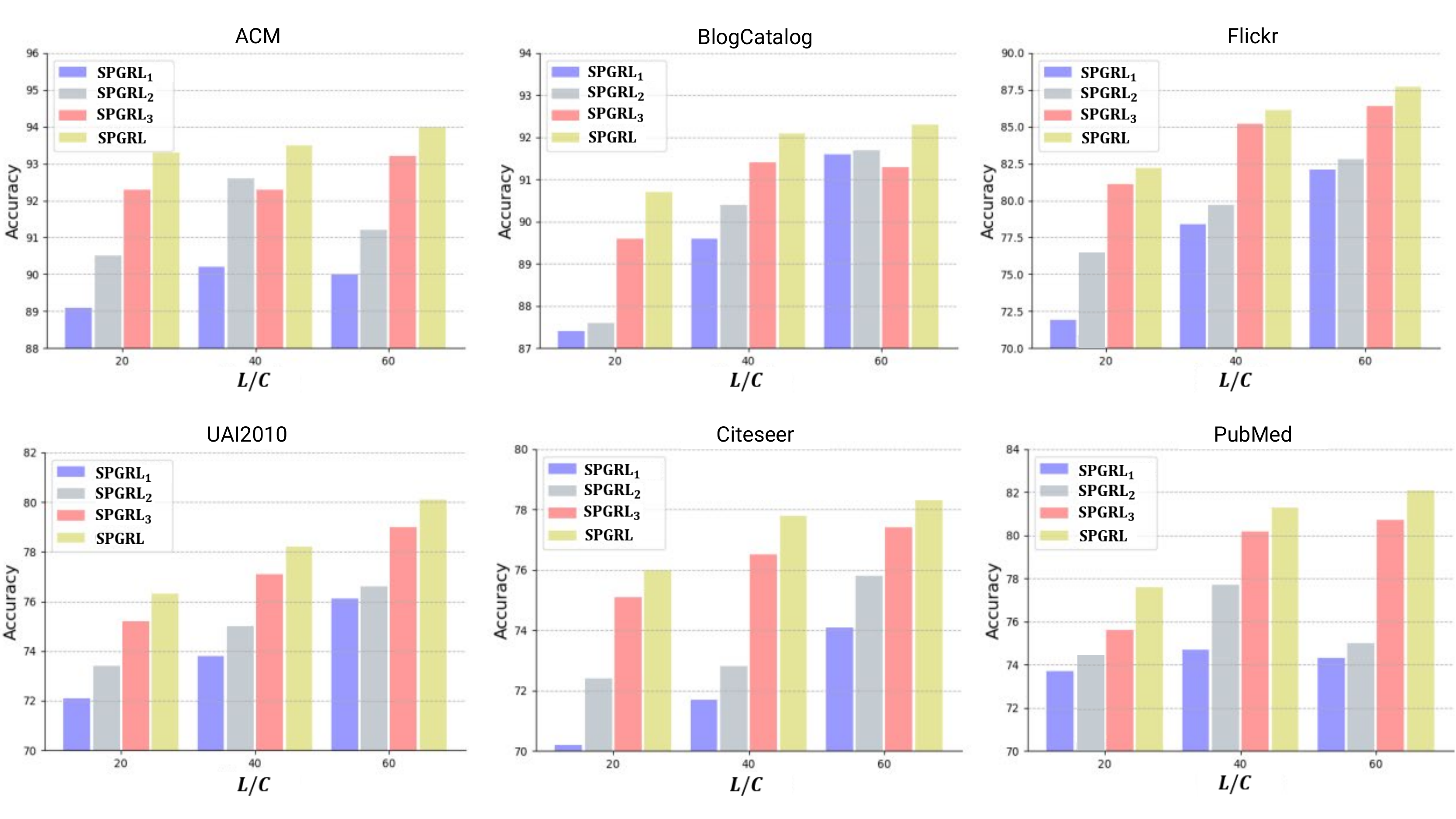}
    \caption{The classification accuracy (\%) of SPGRL and its variants on six datasets. }
    \label{fig: ablation}
\end{figure*}

\subsection{Node Classification Results}
The results of experiments are summarized in Table \ref{table:result}, where the best performance is highlighted in boldface.
Some results are directly taken from~\cite{wang2020gcn,liu2021self}. We have the  following findings:
\begin{itemize}
    \item It can be seen that our proposed method boosts the performance
    of STOA methods across most evaluation metrics on six datasets, which proves 
    its effectiveness. Particularly, compared with other optimal performance, SPGRL achieves a maximum improvement 
    of $4.90\%$ for ACC and $3.84\%$ for F1 on UAI2010. This illustrates that our proposed model can effectively fuse topological structure and feature.
    \item Our SPGRL achieves much better performances than DGI and GMI on all of the metrics. This can be explained by the fact that our method fully exploits the global structure via the MI maximization between graph structure and embedding.
    \item  In most cases, SPGRL produces better performance than SCRL~\cite{liu2021self}, SLAPS~\cite{fatemi2021slaps}, and GCA~\cite{zhu2021graph}, which were published in 2021. This verifies the advantage of our approach.
    \item On some occasions, feature graph produces better result than original graph.
    For example, on BlogCatalog, Flickr, and UAI2010, $k$NN-GCN beats GCN.
    This confirms that incorporating feature graph into our framework can avoid uncertainty or error information in the original graph in many cases.
\end{itemize}

For more intuitive understanding and comparison, 
we use t-SNE method to visualize the progression of the representation learnt by our SPGRL. As shown in Fig.\ref{fig:tsne graph during training}, at the beginning, the representation of ACM is
chaotic and diffused. At epoch 30, a well-learned representation has been established, and the data is divided into different groups. The representations of BlogCatalog and Citeseer evolve in a similar way during training, and they both obtain good representations in 30 epoches.
To further demonstrate the advantage of our proposed method, 
we also visualize the embedding results on Flickr generated by competitive methods GCN, AMGCN, SLAPS, and SCRL, which are shown in Fig.\ref{fig:tsne graph with different methods}. Our SPGRL method produces a compact cluster structure. In other words, our method has the highest intra-class similarity and the most distinct boundaries between different classes.

It is worth pointing out that our MI computation is more efficient than DGI and GMI, which have a high complexity. Specifically, DGI samples the full graph multiple times by readout function and calculates their MI, while GMI maximizes MI between each node and each edge of the original and output graph.
Therefore, both DGI and GMI become computationally inefficient and resource-consuming during training.
To verify the efficiency of SPGRL, we report the averaged training time per epoch when training SPGRL, DGI and GMI in Fig.\ref{fig: time}.
It can be seen that SPGRL always costs much less time than others. For instance, SPGRL costs 49.7ms per epoch but DGI and GMI need 162.8ms and 267.9ms on Citeseer, respectively. Additionally, for larger datasets like PubMed, DGI and GMI are subject to out-of-memory error.

\begin{table}[H]
\caption{Classification accuracy with low label rates.}
\label{table: few_label}
\centering
\resizebox{\linewidth}{!}{
\renewcommand{\arraystretch}{1.0}
\begin{tabular}{c|cccc|ccc}
\hline
Datasets   & \multicolumn{4}{c|}{Citeseer}                                 & \multicolumn{3}{c}{PubMed}                    \\ \hline
L/C & 3 & 6 & 12 & 18 & 2 & 3 & 7 \\ \hline
Label Rate & 0.5\%         & 1\%           & 2\%           & 3\%           & 0.03\%        & 0.05\%        & 0.10\%        \\ \hline
ChebNet~\cite{defferrard2016convolutional}    & 19.7          & 59.3          & 62.1          & 66.8          & 55.9          & 62.5          & 69.5          \\
GCN~\cite{kipf2017semi}       & 33.4          & 46.5          & 62.6          & 66.9          & 61.8          & 68.8          & 71.9          \\
GAT~\cite{velickovic2018graph}       & 45.7          & 64.7          & 69.0          & 69.3          & 65.7          & 69.9          & 72.4          \\
DGI~\cite{veli2018deep}      & 60.7          & 66.9          & 68.1          & 69.8          & 60.2          & 68.4          & 70.7          \\
M3S~\cite{sun2020multi}    & 56.1      & 62.1          & 66.4          & 70.3          & 59.2          & 64.4          & 70.5          \\
GRACE~\cite{Zhu:2020vf}      & 55.4          & 59.3          & 63.4          & 67.8          & 64.4          & 67.5          & 72.3          \\
AMGCN~\cite{wang2020gcn}     & 60.2          & 65.7          & 68.5          & 70.2          & 60.5              &  62.4             & 70.8              \\ 
SCRL~\cite{liu2021self}      & 62.4   & 67.3 & 69.8 & 73.3 & 67.9 & 71.9 & 73.4 \\
GCA~\cite{zhu2021graph}    &62.6	&63.4	&62.7	&60.8	&70.1	&73.2	&75.8      \\ \hline
SPGRL   & \textbf{64.3}   & \textbf{68.4}   & \textbf{71.7}  & \textbf{74.7} & \textbf{70.2} & \textbf{73.4} & \textbf{76.7} \\ \hline
\end{tabular}}
\end{table}

\vspace{-\topsep}\subsection{Ablation Study}

To validate the effectiveness of different components in our model, we compare SPGRL with its three variants on all datasets.
\begin{itemize}
\item \textbf{SPGRL$_1$}: SPGRL without ${L}_{cr}$ and ${L}_{re}$ to show the impact of local and global structure.
\item \textbf{SPGRL$_2$}: SPGRL without ${L}_{re}$ to show the effect of global structure preserving.
\item \textbf{SPGRL$_3$}: SPGRL with traditional reconstruction, i.e., $q_{\phi}(\mathbf{A}|\mathbf{Z^t})$ and $q_{\phi}(\mathbf{\hat{A}}|\mathbf{Z^f})$, to demonstrate the benefit of exchange-reconstruction.
\end{itemize}

According to Fig.\ref{fig: ablation}, we can draw the following conclusions: (1) The results of SPGRL are consistently better than all variants, indicating the rationality of our model. (2) Both local and global structure information are crucial to representation learning. (3) Exchange reconstruction is beneficial by removing some redundant information.


\subsection{Few Labeled Classification}
To further investigate the capability of SPGRL in dealing with scarce supervision data,
we conduct experiments when the number of labeled examples is extremely small.
Taking Citeseer and PubMed for example, we 
select a small set of labeled examples for model training~\cite{li2018deeper}. 
Specifically, for Citeseer, we select 3, 6, 12, 18 nodes per class, corresponding to 
label rates: $0.5\%$, $1\%$, $2\%$, and $3\%$; 
for PubMed, we select 2, 3, 7 nodes per class, corresponding to three label rates:
$0.03\%$, $0.05\%$ and $0.10\%$. To make a fair comparison, we report mean classification accuracy of 10 runs.

From Table \ref{table: few_label}, we can observe that SPGRL outperforms all STOA approaches. For example, SPGRL improves AMGCN, SCRL, GCA by $5.87\%$, $1.91\%$, and $4.40\%$ on average. 
Particularly, the accuracy of GCN, ChebNet, and GAT decline severely when the label rate is very low, 
especially on $0.5\%$ Citeseer, due to insufficient propagation of label information.
By contrast, self-supervised/contrastive approaches are obviously much better because they additionally exploit supervisory signals. 
Though GCA outperforms SPGRL in most cases of Pubmed dataset in Table \ref{table:result}, its performance is worse than our method at low label rate. Thus, fully exploring structure information could alleviate the reliance of label to some extent.

\subsection{ Experiments with Noise Perturbation}
Many recent studies have found that GCN is vulnerable to noise perturbation on node features or graph structure.
  Hence, it is necessary to evaluate the robustness of our method. We perturb node features by injecting independent Gaussian noise. Consequently, our built feature graph is also corrupted. Note that it is computationally expensive to perturb structure and it behaves similarly to feature perturbation to some extent~\cite{xu2019topology}. 
Therefore, there is no need to corrupt original graph structure $\mathbf{A}$ in our setting.
Specifically, we add Gaussian noise to input features: $\mathbf{X} \leftarrow \mathbf{X} + \mathcal{N}(0, \sigma^2)$, where $\sigma$ is the variance of Gaussian noise. 
We compare to a few closely relevant methods, including GFNN~\cite{DBLP:journals/corr/abs-1905-09550}, which employs low-pass filtering to remove noise.

\begin{figure}[!htbp]
\centering
  \includegraphics[width=0.9\linewidth]{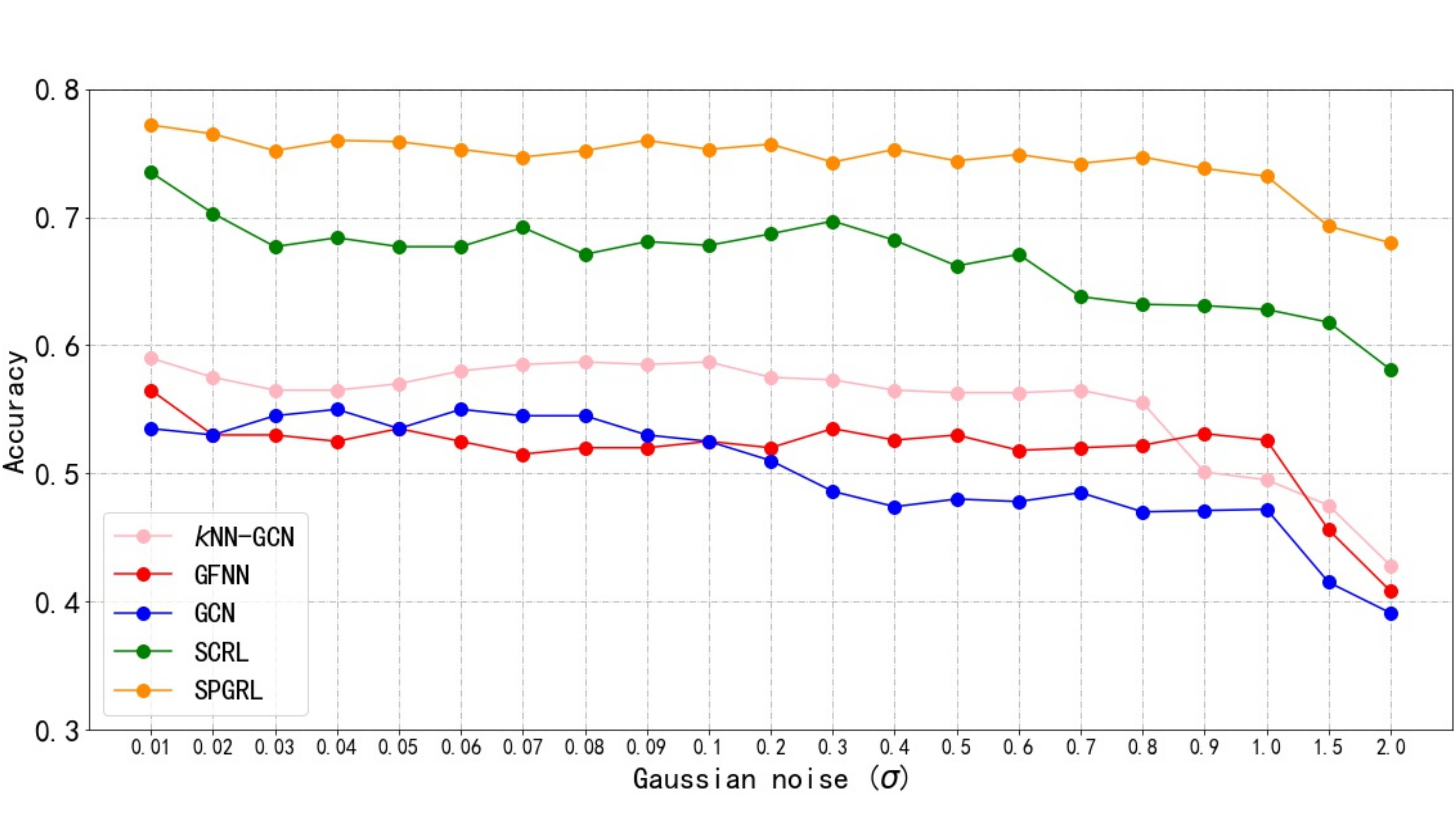}
  \caption{Accuracy of SPGRL under different $\sigma$ on ACM dataset (L/C=20).}
  \label{fig: perturbations}
\end{figure}

 Table \ref{table:noise} shows results with $\sigma=1$ on ACM dataset. We also test with $\sigma\in\{$0.01,0.02,...,2.0$\}$ in Fig.\ref{fig: perturbations}. The results show that SPGRL still performs the best in most scenarios. Its robustness could be explained by the fact that we extract more relevant information from the original graph by maximizing the MI between it and the embeddings, which alleviates the negative influence of noise perturbation.

\begin{table}[H]
\caption{Node classification results with Gaussian noise perturbation ($\sigma = 1.0$).}
\label{table:noise}
\resizebox{\linewidth}{!}{
\renewcommand{\arraystretch}{.99}
\begin{tabular}{c|c|c|c|c|c|c|c}
\hline
Dataset & Metrics & L/C & SPGRL  & SCRL~\cite{liu2021self} &  GFNN ~\cite{DBLP:journals/corr/abs-1905-09550} &\begin{tabular}[c]{@{}c@{}} GCN~\cite{kipf2017semi}  \end{tabular} & $k$NN-GCN~\cite{wang2020gcn}\\ \hline
\multirow{3}{*}{ACM}    & \multirow{3}{*}{ACC}    & 20 & \textbf{73.2} & 62.8 & 52.6 & 47.2 &  49.5\\ \cline{3-8} 
                          &                      & 40 & \textbf{78.1} & 75.2 & 50.1 & 52.1 & 57.4 \\ \cline{3-8} 
                          &                      & 60 & \textbf{86.6} & 80.6 &  56.4 & 57.2 & 58.8 \\  \hline
\multirow{3}{*}{BlogCatalog}& \multirow{3}{*}{ACC} & 20 & \textbf{80.3} & 75.0 & 62.4 & 55.1& 56.1 \\ \cline{3-8}  
                          &                      & 40 & \textbf{86.2} & 77.9 & 47.3 & 57.7  & 63.2  \\ \cline{3-8} 
                          &                      & 60 & \textbf{89.3} & 78.3 & 53.4 & 57.1  & 61.4 \\ \hline
\multirow{3}{*}{UAI2010} & \multirow{3}{*}{ACC}      & 20 & \textbf{72.8}  & 69.4 & 32.8 & 49.9 & 52.0 \\ \cline{3-8}  
                          &                      & 40 & \textbf{73.3} & 73.3  & 32.2 & 53.0 & 55.0\\ \cline{3-8}  
                          &                      & 60 & 76.8 & \textbf{76.9} & 29.5 & 57.5  & 58.9 \\  \hline
\multirow{3}{*}{Flickr} & \multirow{3}{*}{ACC}   & 20 & \textbf{65.3}  & 54.2 & 21.3 & 27.5  & 35.5\\ \cline{3-8} 
                          &                      & 40 & \textbf{65.6} & 61.9 & 20.3 & 31.3  & 29.4 \\ \cline{3-8} 
                          &                      & 60 & \textbf{74.1} & 73.5 & 24.3 & 34.4  & 32.3 \\  \hline
\multirow{3}{*}{Citeseer} & \multirow{3}{*}{ACC} & 20 & 45.3 & \textbf{51.3}  & 36.1 & 34.4  & 36.6\\ \cline{3-8} 
                          &                      & 40 & \textbf{59.8} & 59.7 & 40.9 & 43.4  & 41.6\\ \cline{3-8} 
                          &                      & 60 &\textbf{ 66.2} & 63.2 & 46.0 & 45.7  & 47.8 \\  \hline
\end{tabular}}
\end{table}

\subsection{Parameter Analysis}
\begin{figure}[t]
\centering
\subfigure[ACM]{
		\includegraphics[width=0.21\textwidth]{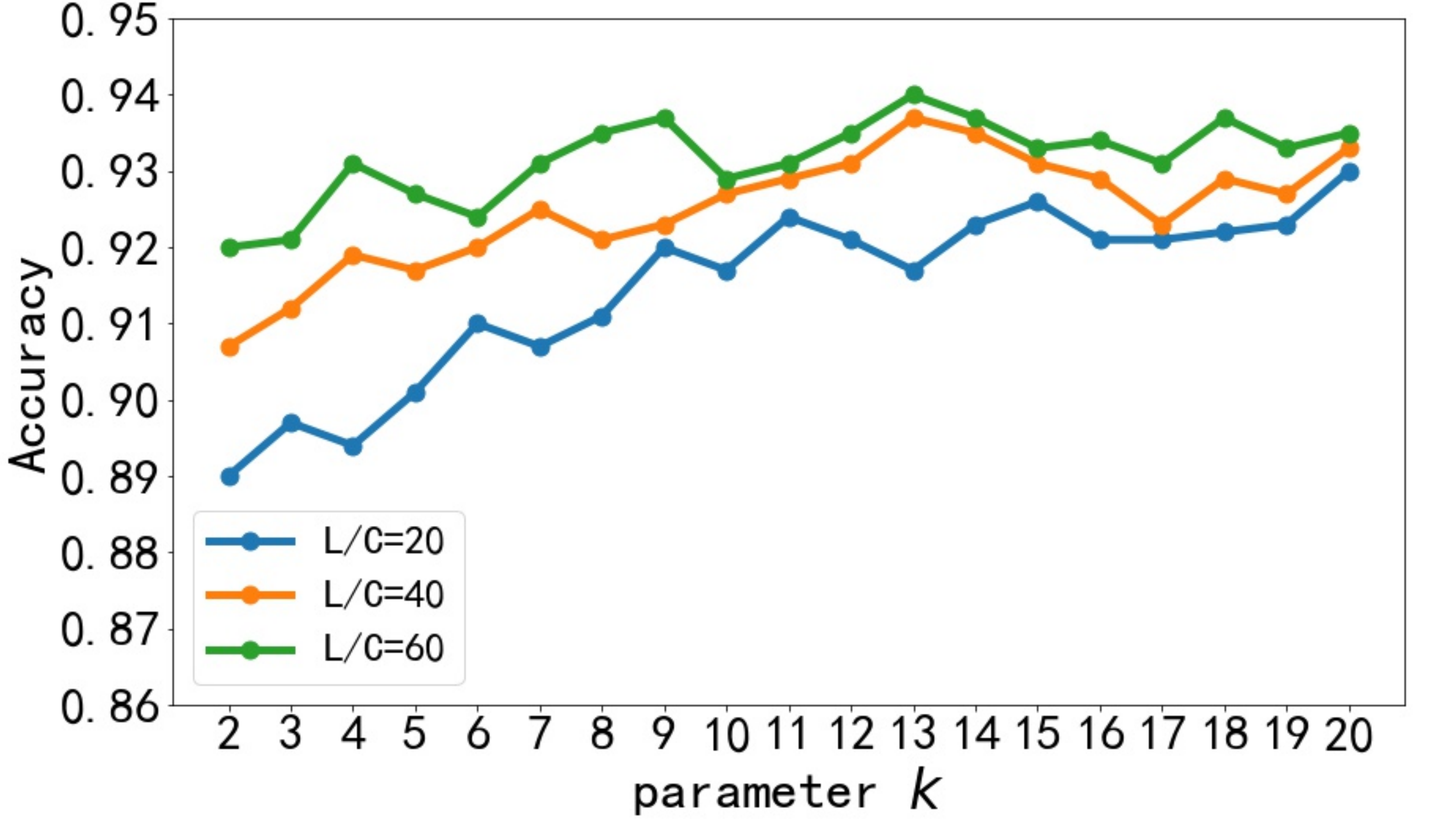}
	}
\subfigure[UAI2010 ]{
		\includegraphics[width=0.21\textwidth]{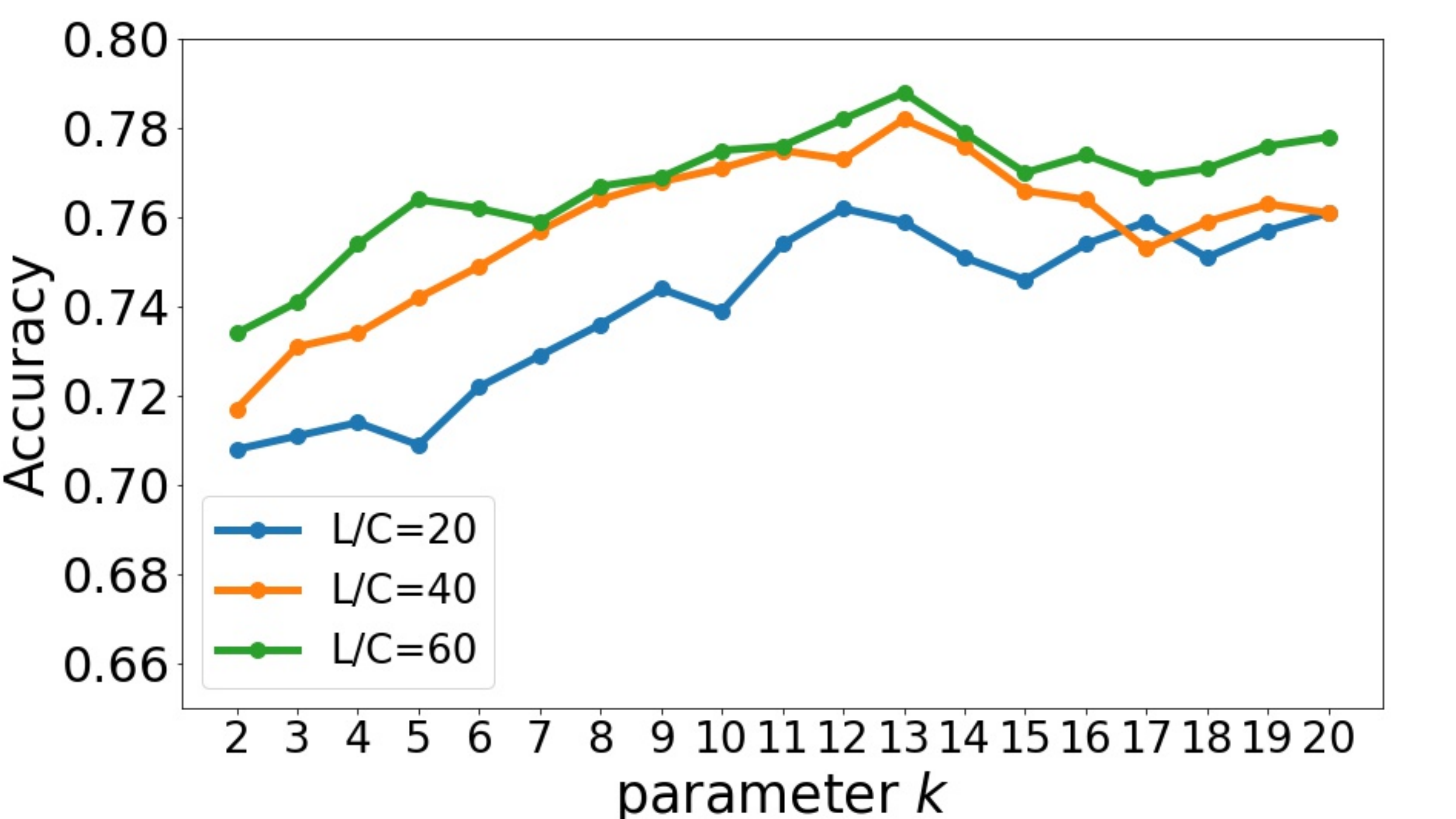}
	}
\caption{The influence of parameter $k$ on ACM and UAI2010 dataset.}
\label{fig:k_acm}
\end{figure}

\begin{figure*}[!t]
\centering
\subfigure[ACM L/C=20]{
		\includegraphics[width=0.27\textwidth]{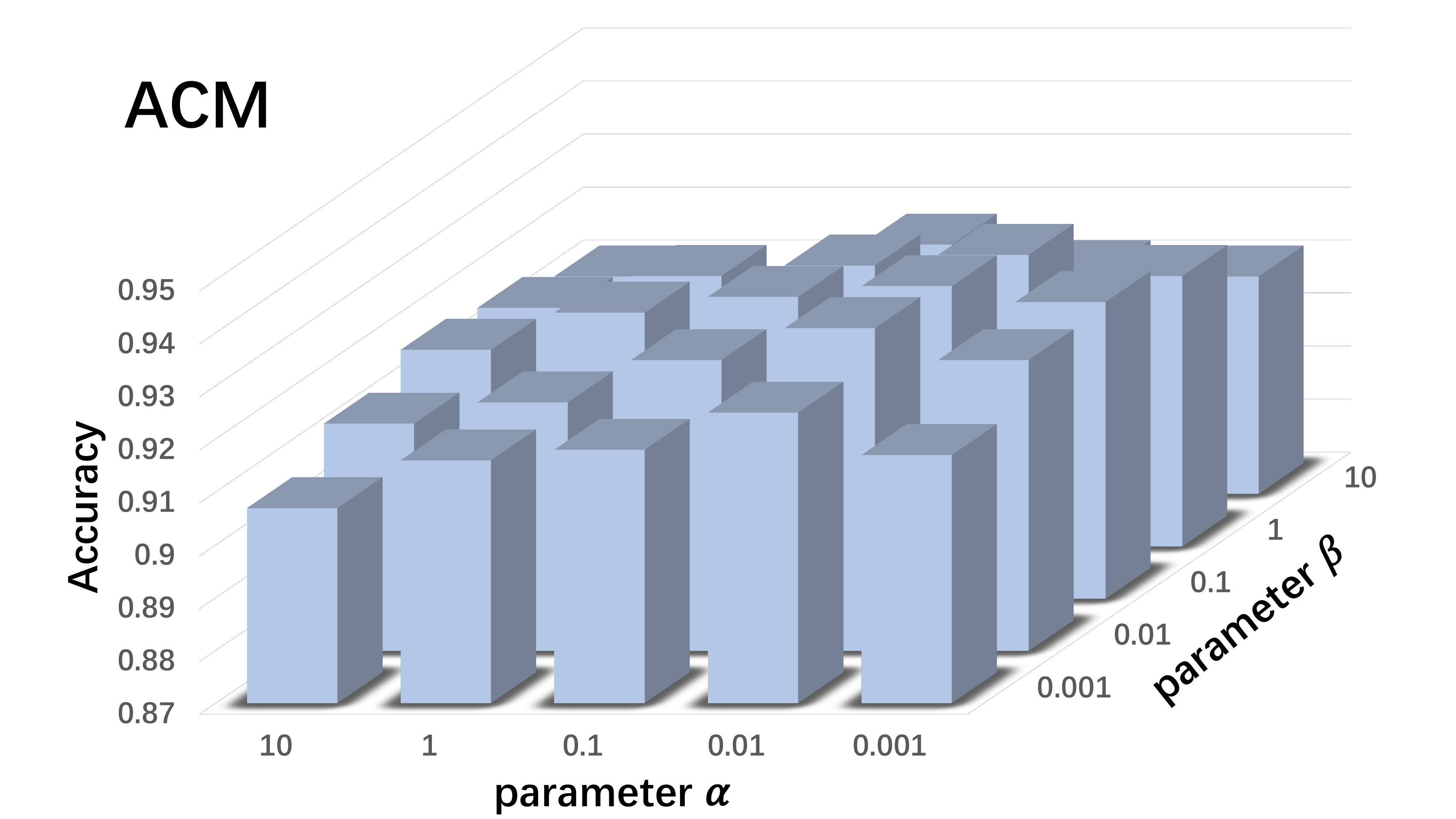}
	}
\subfigure[ACM L/C=40]{
		\includegraphics[width=0.27\textwidth]{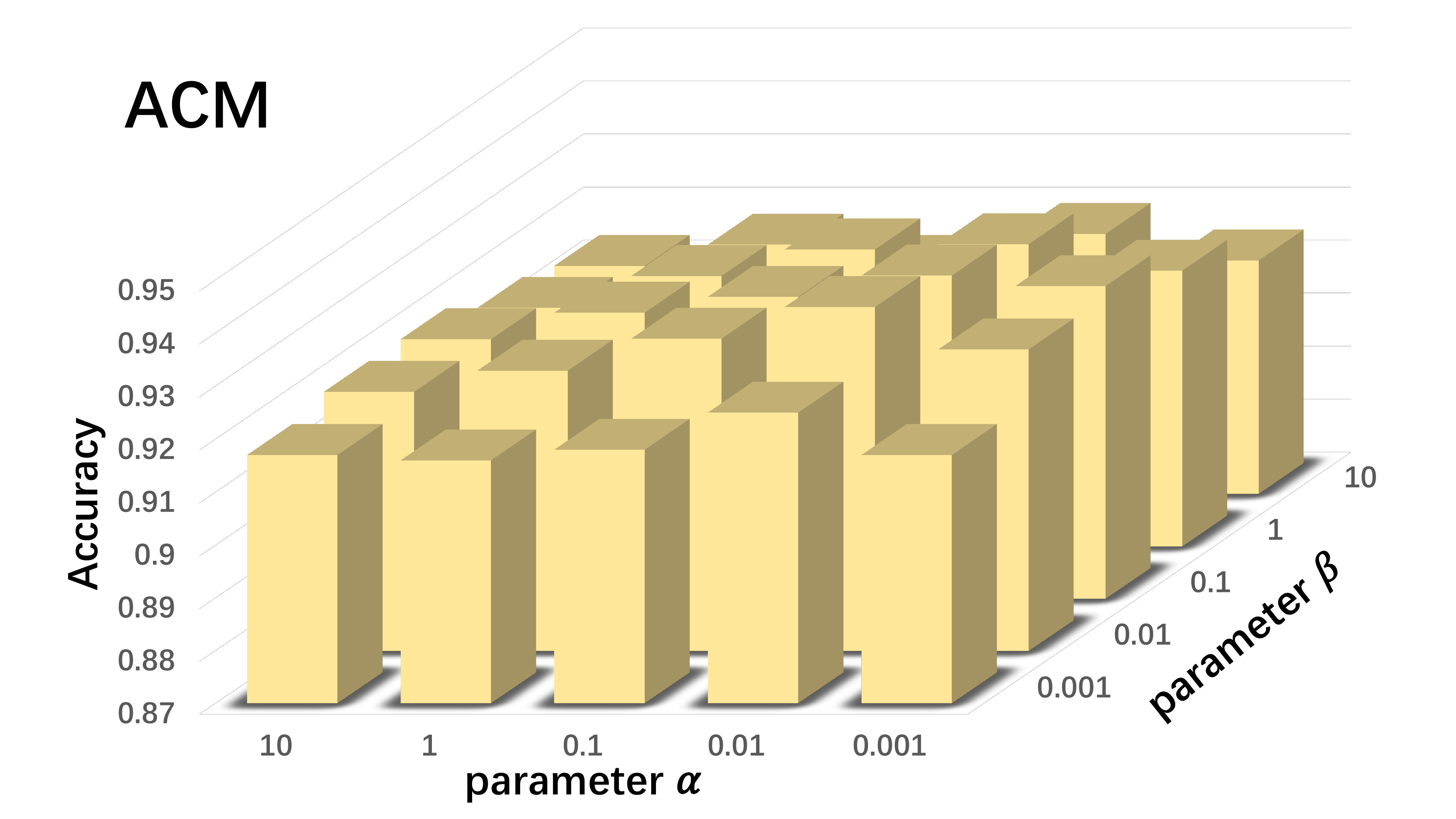}
	}
\subfigure[ACM L/C=60]{
		\includegraphics[width=0.27\textwidth]{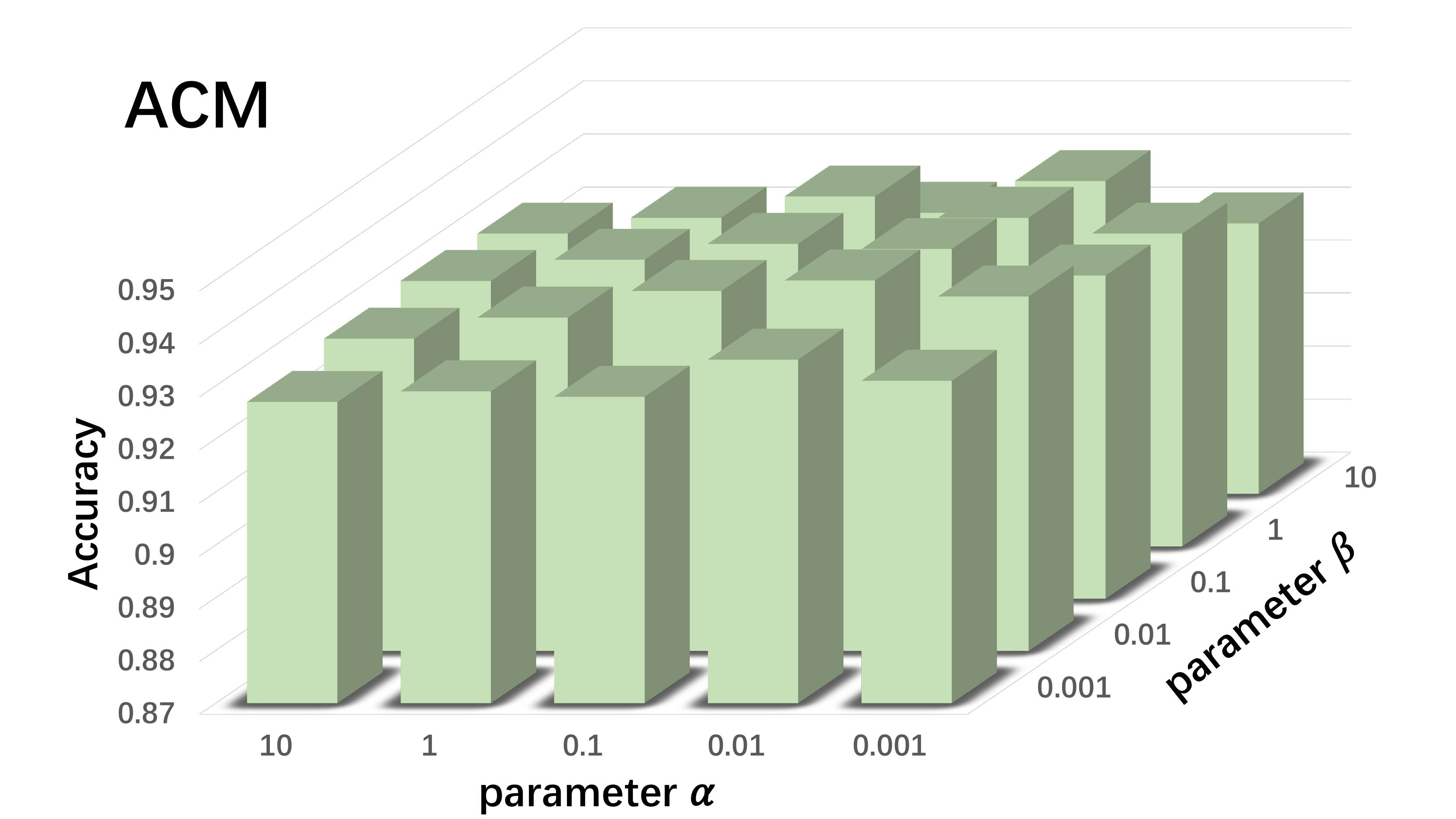}
	}
\subfigure[UAI2010 L/C=20]{
		\includegraphics[width=0.27\textwidth]{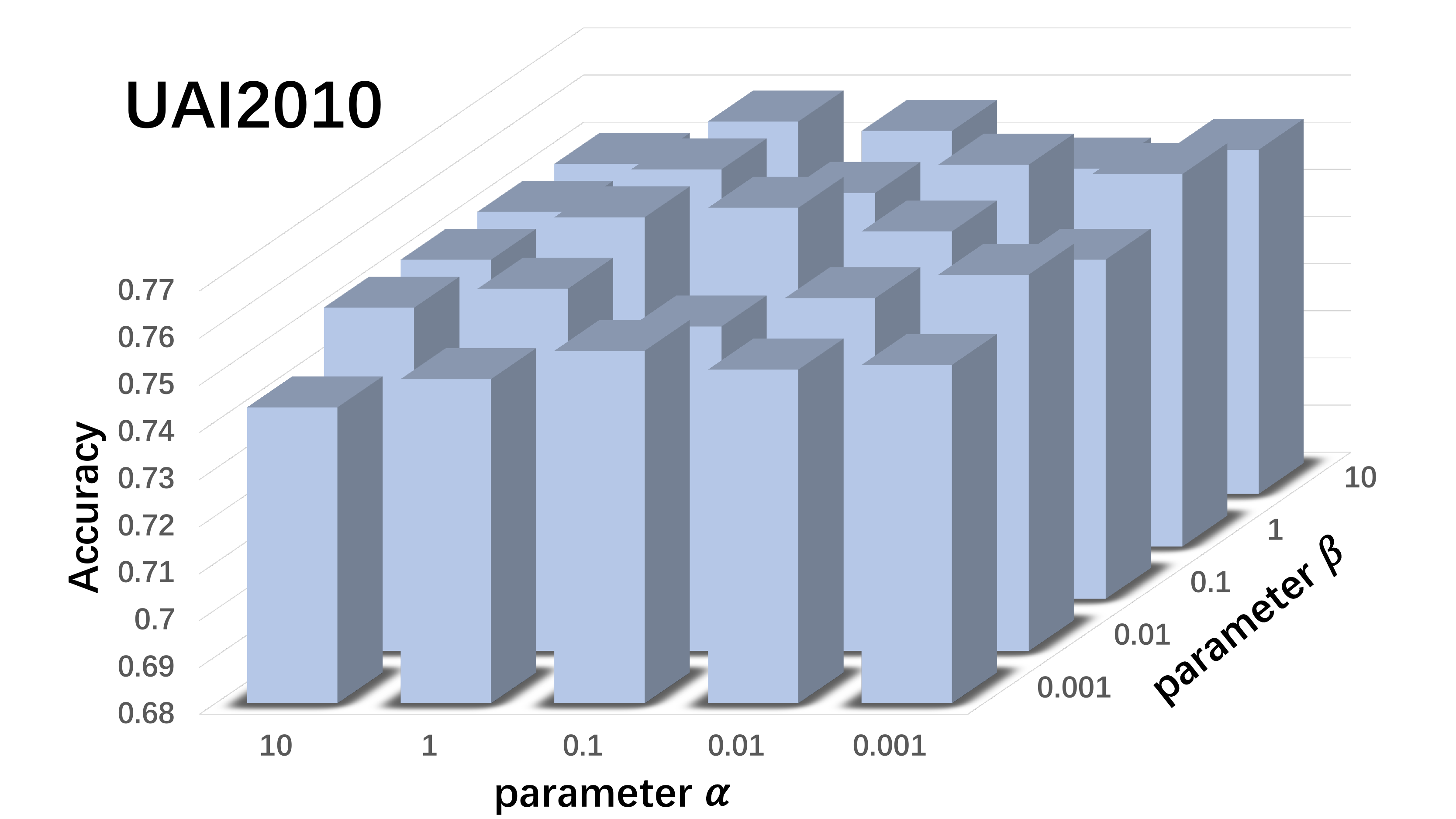}
	}
\subfigure[UAI2010 L/C=40]{
		\includegraphics[width=0.27\textwidth]{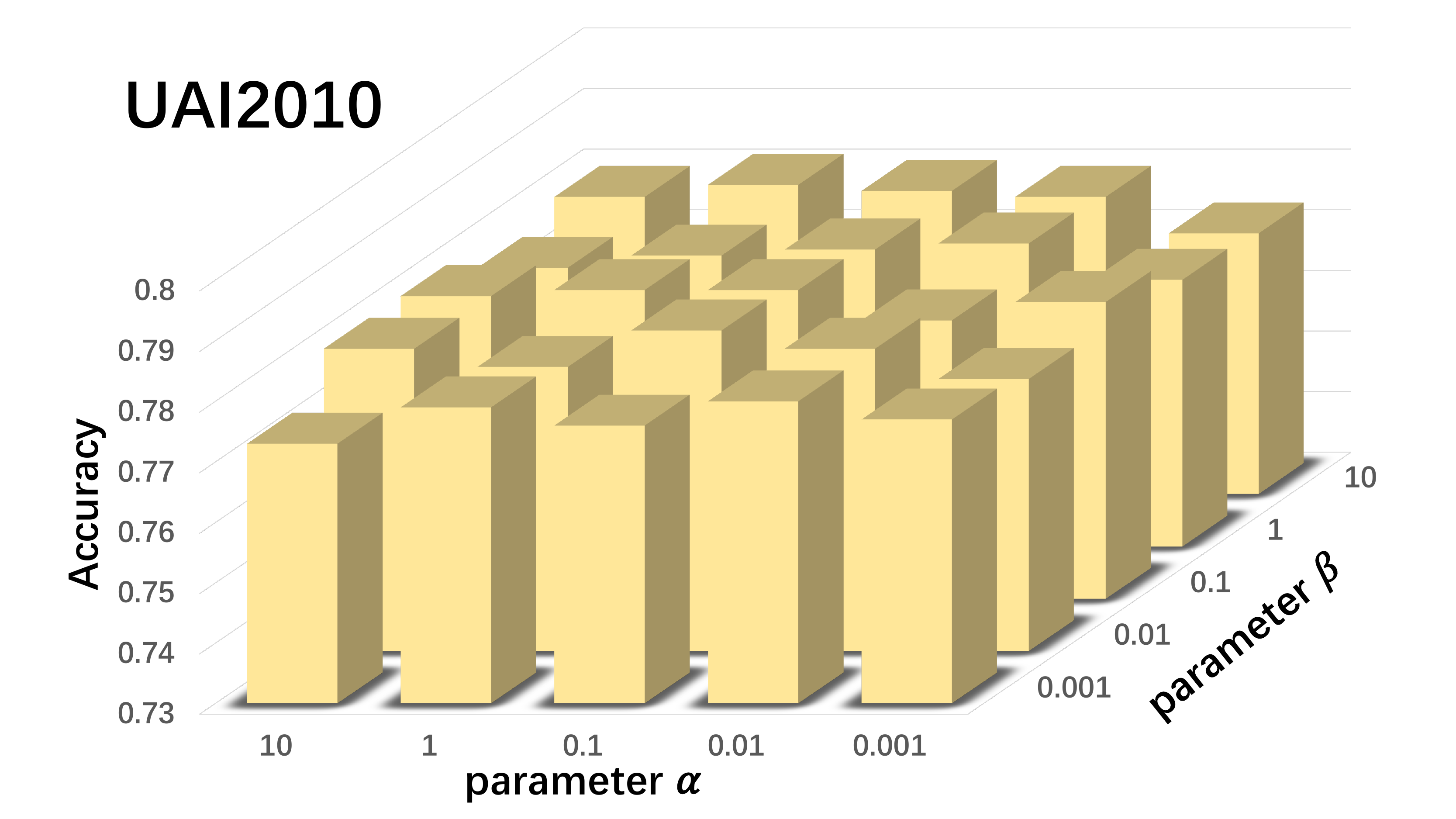}
	}
\subfigure[UAI2010 L/C=60]{
		\includegraphics[width=0.27\textwidth]{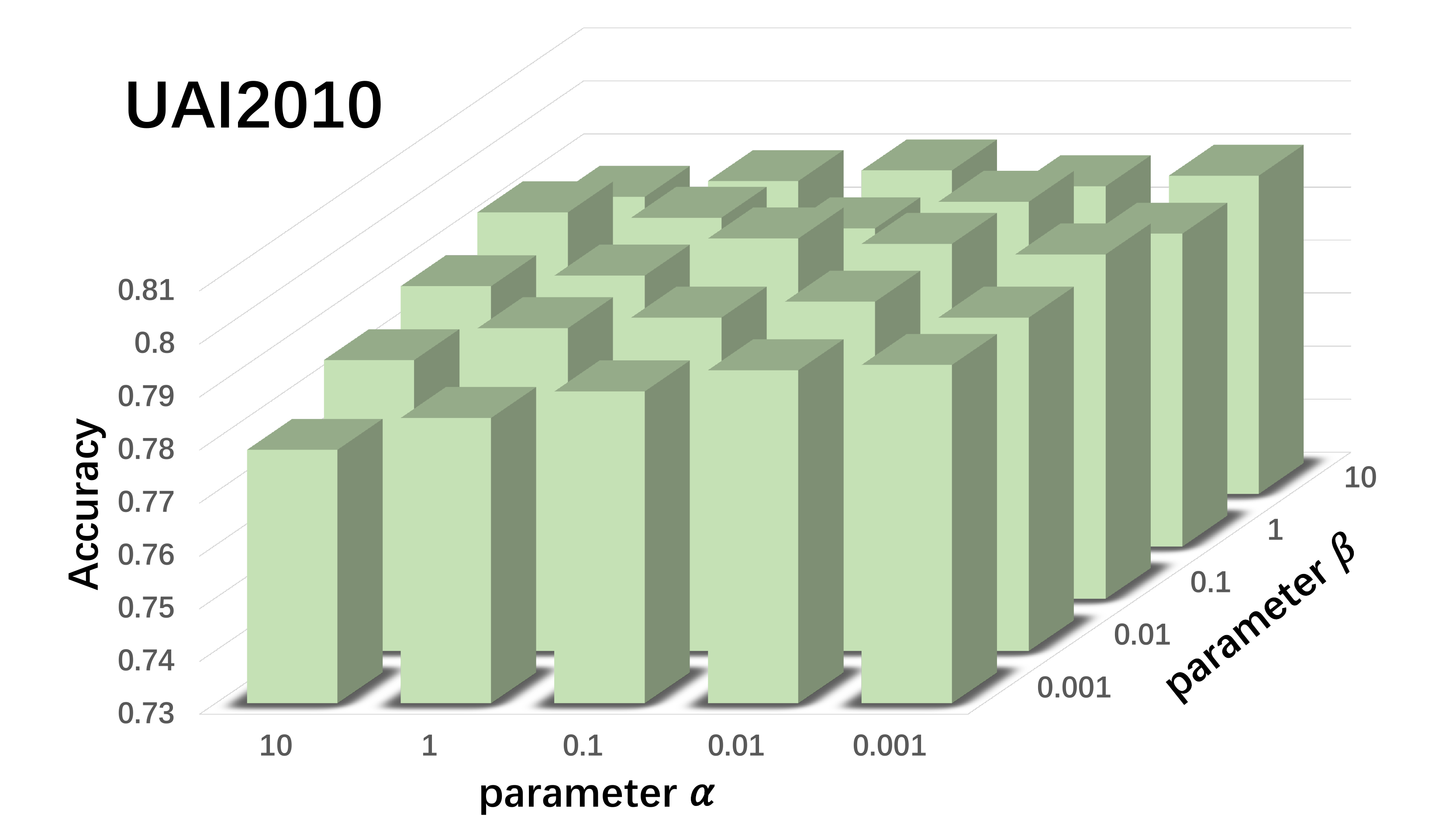}
	}
\caption{The influence of parameters $\alpha$, $\beta$ on ACM and UAI2010 dataset.}
\label{fig:alpha_beta_acm}
\end{figure*}

In this section, we analyze the sensitivity of parameters of our method on ACM and UAI2010 dataset. As shown in Fig.\ref{fig:k_acm},the accuracy usually increases along with $k$. This is reasonable since increasing $k$ means more high-order proximity information is incorporated. On the other hand, extremely large $k$ could also introduce noisy that will deteriorate the performance.
From Fig.\ref{fig:alpha_beta_acm}, we can see SPGRL has competitive performance on a large range of values, which suggests the stability of our method. 

\section{Conclusion}
In this paper, we propose a framework to preserve the local-global structure information during graph embedding. This is mainly realized by maximizing MI between topological structure and feature representation, which is further converted to exchange reconstruction according to our theoretical derivation. Comprehensive experiments verify the effectiveness, efficiency, and robustness of our approach in different scenarios.  

\section{acks}
This work was supported by the Natural Science Foundation of China under Grant 62276053.

\bibliographystyle{IEEEtran}

\bibliography{main}

\end{document}